%% file: main.tex
\documentclass[11pt,a4paper]{article}
\usepackage{times,latexsym}
\usepackage{url}
\usepackage[T1]{fontenc}

\usepackage[acceptedWithA]{tacl2021v1}

\usepackage{xspace,mfirstuc,tabulary}

\usepackage[utf8]{inputenc}

\usepackage{microtype}

\usepackage{inconsolata}

\usepackage{tikz}
\usetikzlibrary{positioning}
\usetikzlibrary{bayesnet}
\usetikzlibrary{arrows}
\usetikzlibrary{backgrounds}
\usepackage{color}
\usepackage{graphicx}

\usepackage{amsmath}
\usepackage{amssymb}

\usepackage{cleveref}

\usepackage{hhline}
\usepackage{colortbl}

\usepackage{tabularx}

\usepackage{relsize}
\usepackage{multirow}

\usepackage{listings}
\usepackage{soul}

\DeclareMathOperator*{\argmax}{arg\,max}

\DeclareMathOperator*{\tp}{tp}
\DeclareMathOperator*{\ibp}{ibp}

\def\fs{\kern 1.35em}

\usepackage{mathtools}

\DeclarePairedDelimiterX{\infdivx}[2]{\big[}{\big ]}{%
  #1\;\delimsize\|\;#2%
}

\title{Hierarchical Indexing for Retrieval-Augmented Opinion Summarization}

\author{{Tom Hosking \qquad Hao Tang \qquad Mirella Lapata} \\
  Institute for Language, Cognition and Computation \\
  School of Informatics, University of Edinburgh \\
  10 Crichton Street, Edinburgh EH8 9AB\\
  \texttt{tom.hosking@ed.ac.uk}\quad \texttt{hao.tang@ed.ac.uk} \quad \texttt{mlap@inf.ed.ac.uk}}

\begin{document}

\include{content}

\end{document}

%% file: content.tex
\maketitle
\begin{abstract}
We propose a method for unsupervised abstractive opinion summarization, that combines the attributability and scalability of extractive approaches with the coherence and fluency of Large Language Models (LLMs). Our method, HIRO, learns an index structure that maps sentences to a path through a semantically organized discrete hierarchy. At inference time, we populate the index and use it to identify and retrieve clusters of sentences containing popular opinions from input reviews. Then, we use a pretrained LLM to generate a readable summary that is grounded in these extracted evidential clusters. The modularity of our approach allows us to evaluate its efficacy at each stage. We show that HIRO learns an encoding space that is more semantically structured than prior work, and generates summaries that are more representative of the opinions in the input reviews. Human evaluation confirms that HIRO generates significantly more coherent, detailed and accurate summaries.
\end{abstract}

\section{Introduction}
\label{sec:intro}

Online review websites are a useful resource when choosing which hotel to visit or which product to buy, but it is impractical for a user to read hundreds of reviews. Automatic opinion summarization aims to aggregate a large and diverse set of customer reviews about a particular \textit{entity} into a single, easy to read summary. A good summary should accurately reflect the balance of opinions in the input reviews, highlighting the most common or \textit{popular} opinions, while omitting unnecessary details. A useful summary should also help \textit{compare} between competing options, and include points that differentiate the current entity from others.

Early work on opinion summarization extracted reviewers' sentiment about specific features \citep{10.1145/1014052.1014073} or identified salient sentences based on centrality \citep{lexrank}, while more recent methods have proposed \emph{extractive} models that use learned feature spaces  \citep{angelidis-etal-2021-extractive,basu-roy-chowdhury-etal-2022-unsupervised}. Prior work on \emph{abstractive} opinion summarization has almost exclusively either required costly supervision \citep{brazinskas-etal-2021-learning,cattan-etal-2023-key} or has assumed that the number of input reviews is limited \citep{coavoux-etal-2019-unsupervised,brazinskas-etal-2020-unsupervised,amplayo-etal-2021-aspect, amplayo2021unsupervised,iso-etal-2021-convex-aggregation}. This defeats the point of a summary: a user could feasibly read 8 reviews in a reasonable period of time. A good summarization system should be \textit{scalable}, since popular products online may receive thousands of reviews. It should also be \textit{attributable}, offering some evidence to justify its output. Paraphrasing \citet{10.1162/coli_a_00486}, we say that a statement $s$ is attributable to some evidence $E$, if a generic reader would agree that `According to $E$, $s$ is true'. Finally, it should generate summaries that are \textit{coherent} and \textit{faithful} to the input reviews.

Large Language Models (LLMs) have been shown to generate highly fluent summaries in the news domain \citep{bhaskar-etal-2023-prompted} but are a flawed solution because current instruction-tuned models are not attributable, and because their context windows limit the number of reviews they are able to base their summaries on. Models with long context windows have been proposed \citep{Beltagy2020Longformer,s4} but these are not currently instruction-tuned, and it has been shown that LLMs are biased toward information at the start and end of the input \citep{liu2023lost}.

Our approach, \textbf{H}ierarchical \textbf{I}ndexing for \textbf{R}etrieval-Augmented \textbf{O}pinion Summarization (HIRO), identifies informative sentences using hierarchical indexing and then passes the selected sentences as input to a LLM, similar to retrieval-augmented generation \cite[RAG,][]{rag}. By separating the steps of content selection and generation, we can combine the attributability and scalability of the discrete representation with the strong generative abilities of LLMs, leading both to a higher quality index and to more informative and coherent output summaries. 

HIRO consists of three modules, allowing for increased control, flexibility and interpretability. The \textbf{Hierarchical Indexer} is an encoder that maps sentences from reviews to paths through a hierarchical discrete latent space. The \textbf{Retriever} uses the index to identify clusters of sentences for each entity that contain popular and informative opinions. These sentence clusters are passed to a \textbf{Generator}, a pretrained LLM, that generates coherent summaries that are grounded in the retrieved sentences.

Our contributions are as follows:
\begin{itemize}
  \setlength\itemsep{-0.2em}
  \vspace{-0.2cm}
  \item We propose a method for learning an encoder that maps sentences to a path through a semantically structured discrete hierarchy.
  \item We show how to exploit this discrete hierarchy at inference time to identify clusters of related and prevalent sentences from input reviews.
  \item We introduce an automatic metric that measures whether generated summaries reflect the input reviews, while penalizing overly generic statements.
  \item Through extensive experiments on two English datasets from different product domains, we demonstrate that passing these retrieved sentences in a zero-shot manner to a pretrained LLM generates summaries that better reflect the distribution of opinions within the input reviews. Human evaluation shows that summaries generated by HIRO are significantly more coherent and accurate than prior work, and are preferred by annotators.
  \vspace{-0.2cm}
\end{itemize}
Our code and dataset splits are available at \mbox{\url{https://github.com/tomhosking/hiro}}.

\section{Related Work}
\label{sec:relatedwork}

\paragraph{Opinion Summarization} Prior approaches to generating summaries of reviews broadly fall into two categories. \textit{Extractive} methods generate summaries by selecting representative sentences from input reviews to use as the summary \citep{lexrank,angelidis-etal-2021-extractive,basu-roy-chowdhury-etal-2022-unsupervised}. These types of approach are scalable and inherently attributable, but result in summaries that are overly detailed and not coherent. \textit{Abstractive} methods `read' input reviews and generate novel language to use as the summary  \citep{brazinskas-etal-2020-unsupervised,iso-etal-2021-convex-aggregation}. The resulting summaries are more fluent and coherent, but most prior abstractive methods are only able to consider a limited number of input reviews, whether because of combinatoric complexity or a maximum input length.

\citet{hosking-etal-2023-attributable} propose a hybrid method that represents sentences from reviews as paths through a learned discrete hierarchy, then generates output sentences based on frequent paths through this hierarchy. While their method is abstractive as well as attributable and scalable, the highly compressed bottleneck leads to frequent hallucination and generic output with poor coherence.


\citet{louis-maynez-2023-opinesum} use a Natural Language Inference (NLI) model to construct `silver standard' summaries to use as training data for fine-tuning a pretrained language model \cite[T5,][]{t5}. However, their approach is computationally very expensive, calculating over 1B pairwise entailment relations between sentences in the training data, before fine-tuning a LLM. By contrast, HIRO uses a lightweight indexing encoder, combined with an off-the-shelf LLM that is prompted in a zero-shot manner.


Other work has used specialised training data to train supervised models \citep[e.g.,][]{10.1145/3477495.3532037,cattan-etal-2023-key}; however, such data is expensive to collect and is not generally available in every language, domain, or setting.

\paragraph{Structured Encodings and Indexes}

Prior work has investigated how to learn representations of data with richer structure and interpretability. \citet{DBLP:journals/corr/VendrovKFU15} propose learning an embedding space where the ordering of two pairs of samples could be inferred from their relative positions, but their method requires supervision of the correct ordering. \citet{opper-etal-2023-strae} describe a method for learning embeddings that explicitly include structural information, but they focus on representing structures, rather than on learning an ordering within the embedding space itself. \citet{10.1145/3539618.3591651} learn a tree-based index for passage retrieval concurrently with the dense embedding space, showing that this leads to improved retrieval performance. However, they focus on retrieving passages relevant to a query, rather than aggregating opinions. \citet{sarthi2024raptor} use a pretrained embedding model to cluster related sentences then generate a summary using an LLM, and repeat this process iteratively to construct a tree-structured index over inputs. However, this process is expensive and must be repeated for each new set of inputs.


\paragraph{Content Selection}

The idea of first selecting relevant parts of an input before generating output has been well studied \cite[][\textit{inter alia}]{kedzie-etal-2018-content,DBLP:conf/aaai/Puduppully0L19,amplayo2021unsupervised,narayan-etal-2023-conditional}, and has been shown to be very effective when used in conjunction with LLMs in the form of retrieval-augmented generation \cite[RAG,][]{rag}. \citet{xu2023retrieval} found that retrieval augmentation is beneficial even when using models that can accept long inputs. \citet{wang2023learning} show that including an additional filtering or selection step to RAG is better than naively passing all retrieved documents as input.

\paragraph{Evaluation of Summaries}

Automatic evaluation of generated summaries is extremely challenging. Prior work has shown that ROUGE  \citep{lin-2004-rouge} scores correlate poorly with human assessments of summary quality \citep{callison-burch-etal-2006-evaluating,tay-etal-2019-red,10.1162/tacl_a_00373,shen2023opinsummeval,clark-etal-2023-seahorse,aharoni-etal-2023-multilingual}. Some datasets are created automatically, with references that are not directly based on the input reviews \citep{brazinskas-etal-2021-learning}. Modern summarization system outputs are now often preferred to human-written references \citep{goyal2022zeroshotnews,bhaskar-etal-2023-prompted}.


SummaC \citep{laban-etal-2022-summac} is a \textit{reference-free} metric that uses an NLI model to evaluate the degree of support between a summary and the input documents, but it overly rewards trivial statements; using the obvious statement ``The hotel was a building'' as a summary for every entity achieves a near-perfect SummaC score of 99.9\% on \textsc{Space}, a dataset of hotel reviews \citep{angelidis-etal-2021-extractive}.

\citet{prevalence} propose a metric that uses a NLI model to evaluate \textit{prevalence}, i.e. how many input reviews contain supporting evidence for each sentence in a summary, and explicitly penalizes trivial or redundant output. However, we found it has a similar failure mode to SummaC, with the statement ``The rooms are clean and comfortable'' achieving a prevalence score of 72\% on \textsc{Space}. We propose a modification to prevalence in \Cref{sec:part3_autoeval} that penalizes overly generic summaries.








\begin{figure}[t!]
    \centering
    \includegraphics[width=0.42\textwidth]{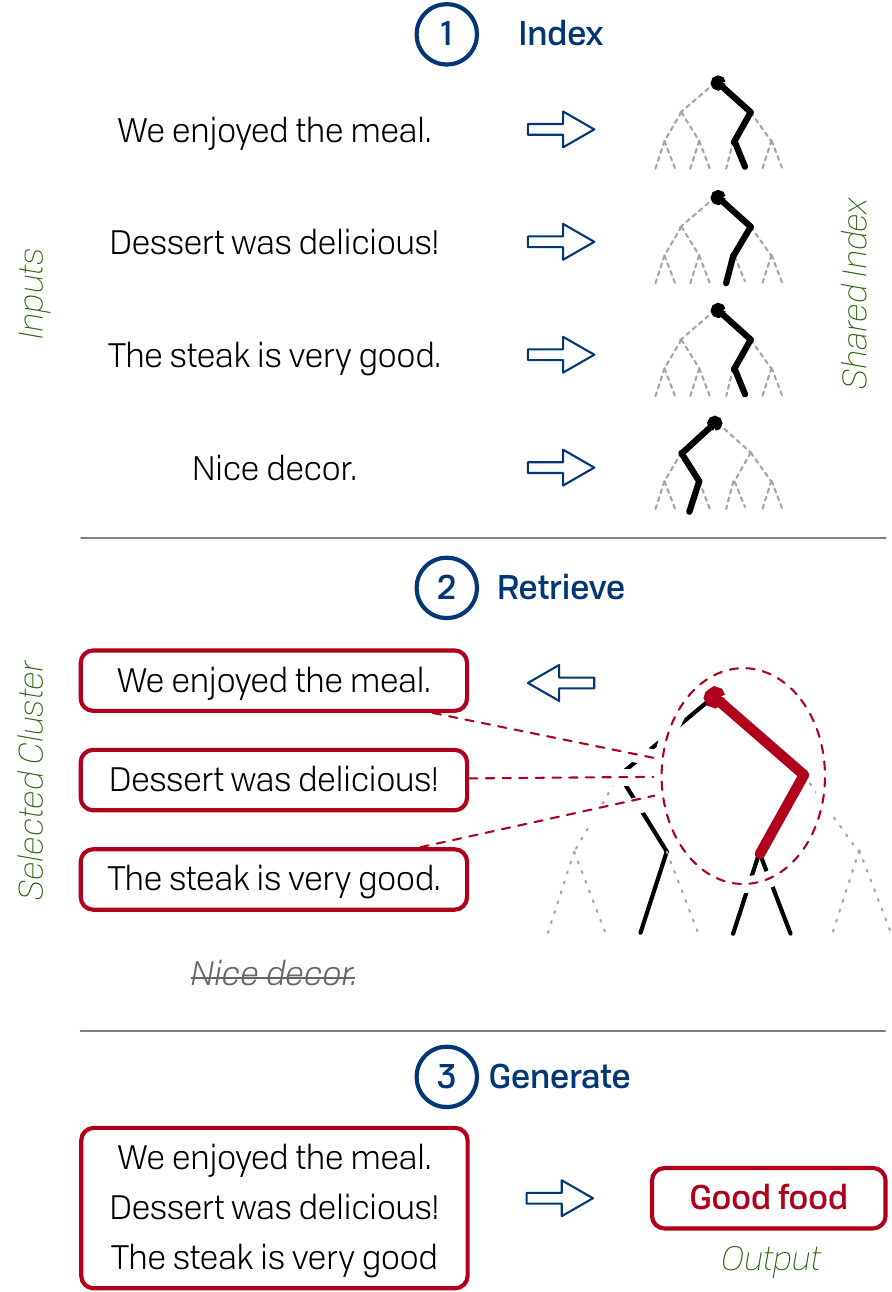}  
    \vspace{-0.2cm}
    \caption{HIRO uses three modules to generate summaries of customer reviews. First, we use our encoder to \textbf{index} all sentences from input review into a learned hierarchy. Then we identify paths within this index that occur frequently, and \textbf{retrieve} the corresponding clusters of sentences. Finally, we pass these clusters to an LLM to \textbf{generate} an output summary.}
    \vspace{-0.2cm}
    \label{fig:pipeline}
\end{figure}

\section{Overview}

Let $\mathcal{R}_e$ be a set of reviews about an entity $e \in \mathcal{E}$, where each review $\mathbf{R} \in \mathcal{R}_e$ is composed of a number of sentences $\mathbf{x}$. The goal is to generate a textual summary that includes the most informative opinions from $\mathcal{R}_e$, while abstracting away the details specific to any one review.

HIRO generates summaries following a modular approach, depicted in \Cref{fig:pipeline}. We learn an index structure that maps each sentence $\mathbf{x}$ from the input reviews to a path $q_{1:D}$ through a discrete hierarchy. We choose a hierarchical representation so that sentences are grouped at a useful level of abstraction; the upper levels of the hierarchy should partition sentences by topic, while the lowest levels should group together sentences with equivalent meaning.

At inference time, we encode all sentences from the reviews, then identify the paths or \textit{subpaths} $q_{1:d}$ within the index that are particularly \textit{popular}, and retrieve the corresponding sentences. This selection process is query-free, instead relying on properties of the hierarchical index to determine the frequency of different opinions. By indexing sentences hierarchically according to their semantics, we can easily identify opinions or topics that occur frequently by simply counting their occurrence in the index.

Finally, we generate a summary by passing the retrieved clusters of sentences as input to a LLM. This \textit{retrieval-augmented} usage of LLMs allows us to benefit from the fluency and coherence of LLMs, while retaining the attributability and scalability of extractive opinion summarization methods. 

We now detail the three modules that comprise HIRO, evaluating each of them in turn to confirm their efficacy compared to previous methods.

\section{Learning a Hierarchical Indexing}
\label{sec:part1}

Our goal is to learn an encoding space where sentences with similar meanings are grouped together. The space should be \textit{discretized} so that frequent opinions can be easily identified by counting the membership of each part of the index, and it should be \textit{hierarchical} so that opinions may be aggregated at an appropriate level of granularity, rather than by details or phrasings specific to a particular review. Finally, the encoding space should be structured semantically, to enable accurate aggregation of opinions; sentences with equivalent meaning should clearly be indexed to the same point in the hierarchy, while sentences that are topically related but not equivalent should be grouped together at a higher level.


We base our encoder on Hercules \citep{hosking-etal-2023-attributable}. Hercules uses an encoder-decoder architecture with a discrete hierarchical bottleneck to generate summaries. It is trained as a denoising autoencoder, and therefore needs to learn a representation that is both compressed enough to enable aggregation, but also expressive enough for the decoder to be able to generate meaningful output. These factors are in direct competition, with the compressed bottleneck leading to output that is generic and contains hallucinations. By contrast, HIRO uses an external LLM as the `decoder', allowing us to focus on learning a representation that is useful for identifying informative opinions. 


\subsection{Method}

The HIRO encoder module maps a single sentence~$\mathbf{x}$ to a path $q_{1:D}$ through a discrete hierarchy, using a technique based on residual vector quantization \citep{rvq,10.1109/TASLP.2021.3129994,hosking-etal-2023-attributable}. 

First, we use a Transformer encoder followed by attention pooling \citep{Vaswani2017,liu-lapata-2019-hierarchical} to map a sequence of tokens $\mathbf{x}$ to a single dense embedding $\mathbf{z} \in \mathbb{R}^{\mathbb{D}}$. Then, we decompose $\mathbf{z}$ into a path through a latent discrete hierarchy $q_{1:D}$, where $q_d \in [1, K]$ are discrete `codes' at each level $d$. Briefly, we induce a distribution over codes at each level~$p(q_d)$, parameterised by a softmax with scores~$s_d$ given by the Euclidean distance from learned codebook embeddings to the residual error between the input and the cumulative embedding from all previous levels, 
\begin{align}
\vspace{-.2cm}
\hspace{-0.25cm} s_d(q) = - \left( \left[\mathbf{z} - \sum\limits_{d'=1}^{d-1} \mathbf{C}_{d'}(q_{d'}) \right ] - \mathbf{C}_d(q)  \right ) ^2\hspace{-0.25cm} ,
\vspace{-.2cm}
\end{align}
where $\textbf{C}_d \in \mathbb{R}^{K \times \mathbb{D}}$ is a codebook which maps each discrete code to a continuous embedding $\textbf{C}_d(q_d) \in \mathbb{R}^{\mathbb{D}}$. During training, we use the Gumbel reparameterization
\citep{jang2016categorical,maddison2017concrete,sonderby2017continuous}
to sample from the distribution~$p(q_d)$. During inference, we set $q_d = \argmax s_d$.

Since our goal is to learn a representation where semantically similar sentences are grouped together, we use a training objective that explicitly induces this arrangement in encoding space. We train the encoder with a contrastive learning objective, bringing representations of semantically similar sentences (i.e., positive pairs) together, and pushing dissimilar ones apart. 

\begin{figure}[t!]
    \centering
    \includegraphics[width=0.48\textwidth]{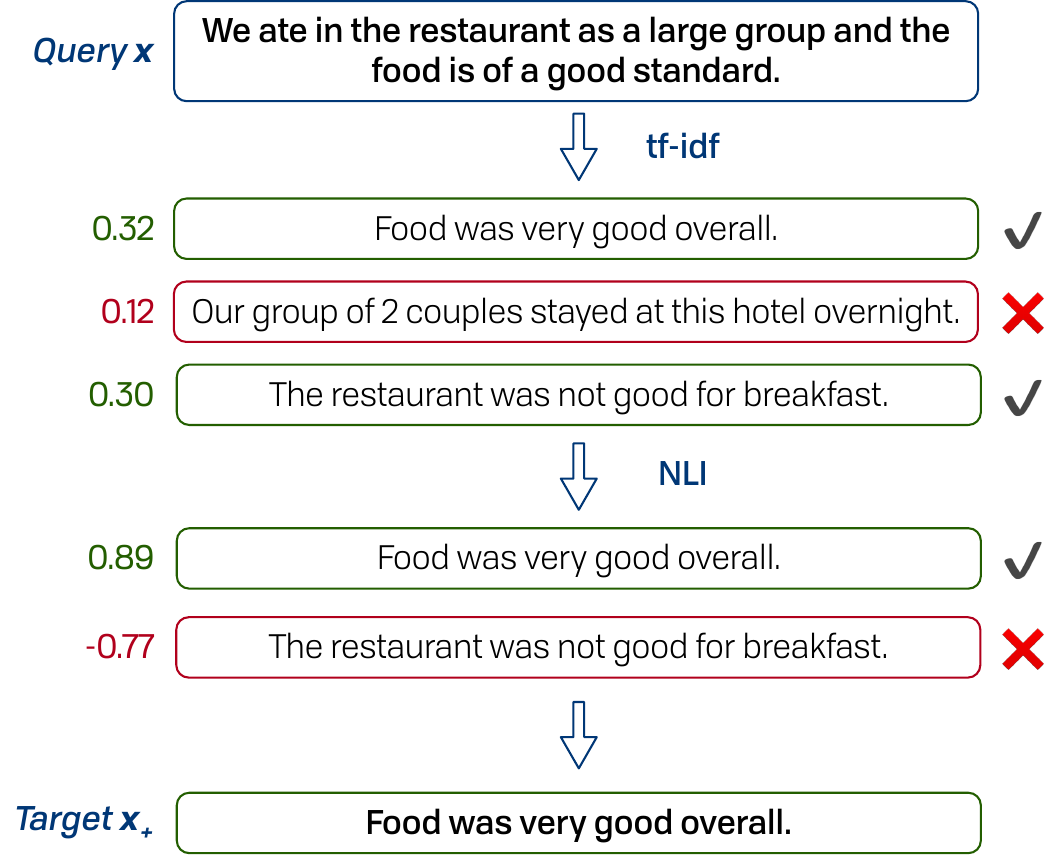}  

    \vspace{-.2cm}
    \caption{An example of the process for constructing the positive pairs used to train our model. Given a query sentence, we first use \mbox{tf-idf} to identify possible candidates from the training data, keeping only those sentences with similarity over a specified threshold. Then we check for entailment using an NLI model, and use any sentences labelled as `entailed' as positive targets. }
    \vspace{-.2cm}
    \label{fig:trainingdata}
\end{figure}

For each sentence in the training data, we construct positive pairs of semantically related sentences~$\mathbf{x},\mathbf{x}_+$ as follows: given a random `query' sentence $\mathbf{x}$ from the training data, we identify possible candidate `targets' based on tf-idf similarity; then, we check whether each candidate is entailed by the query using an NLI model \cite[DeBERTa v3, trained on Debiased NLI;][]{he2021debertav3,wu-etal-2022-generating}, and use the sentences which are labelled as entailed as positive targets~$\mathbf{x}_+$. An example is shown in \Cref{fig:trainingdata}. We do not use any `hard' negatives; instead, during training we calculate the pairwise tf-idf similarity between all samples in a batch, and include only those samples with similarity below a threshold as negatives $\mathbf{X}_{-}$. We set this threshold to $0.3$ based on validation set performance. This prevents `false negatives' being wrongly forced apart, and allows sentences that are topically related but not strictly equivalent to remain in the same high-level grouping within the index.

We found that it was crucial to include sufficient exploration in the process of constructing positive pairs. The candidate sentences retrieved using \mbox{tf-idf} should be sufficiently similar that we are likely to find ones that are entailed, but not so similar that they only have minor lexical differences.

We use a modified version of the InfoNCE training objective \citep{infonce} designed to bring the representations of a query $\mathbf{x}$ closer to those of a positive target $\mathbf{x}_+$, while pushing them apart from negative samples $\mathbf{X}_{-}$,
\begin{align}
    \mathcal{L} =& - \rho(\mathbf{x},\mathbf{x}_+) \log f,\\f=&  \frac{\exp\big(s(\mathbf{x},\mathbf{x}_+)\big)}{ \exp\big(s(\mathbf{x},\mathbf{x}_+)\big) + \frac{\omega}{\vert\mathbf{X}_{-}\vert} \sum\limits_{\mathbf{x}_{-}\in \mathbf{X}_{-}} \hspace{-0.8em}\exp\big(s(\mathbf{x},\mathbf{x}_{-})\big) }, \notag
\end{align} 
where $\rho(\mathbf{x},\mathbf{x}_+)$ is the tf-idf similarity between $\mathbf{x}$ and $\mathbf{x}_+$ that weights the \textit{confidence} of the positive pairs, inspired by MarginMSE \citep{hofstatter2021improving}, and $\omega$ is a constant that controls the strength of the negative examples. The similarity function $s(\cdot,\cdot)$ is given by the mean dot product between the embedding of all subpaths $q_{1:d}$ for $d\leq D$,
\begin{align}
    \hspace{-0.2cm}s(\mathbf{x},\mathbf{x}') = \frac{1}{D} \sum\limits_{d=1}^D \max\big(  \mathbf{C}(q_{1:d})^T \mathbf{C}(q'_{1:d}), 0\big),
\end{align}
where $\mathbf{C}(q_{1:d}) = \sum_d \mathbf{C}_d(q_{d})$ is the full embedding of path $q_{1:d}$. Intuitively, this brings together the representations of the positive pairs at each level in the hierarchy, while penalizing any overlap with the representations of negative examples.

The similarity is floored at zero, to prevent the model from being able to `game' the loss by pushing negative examples further and further apart. Although the positive pairs are selected based on a directional entailment label, we do not exploit this directionality in our training objective.

We employ the techniques to induce a hierarchical encoding space proposed by \citet{hosking-etal-2023-attributable}, including depth dropout, initialization decay, and norm loss. We additionally include the entropy of the distribution over codes, $-\sum_{d,q_d}\log\big(p(q_{d})\big)$, as an additional term in the objective, to encourage exploration of the latent space during training.

\subsection{Evaluation}

We now evaluate whether the combination of discrete hierarchical encoding and contrastive objective leads to a more semantically distributed representation than previous methods.

\paragraph{Experimental Setup}
\label{sec:part1_eval}

We experiment using \textsc{\textbf{Space}} \citep{angelidis-etal-2021-extractive}, which consists of hotel reviews from TripAdvisor with 100 reviews per entity, as well as reference summaries created by annotators. We encode all the review sentences from the \textsc{Space} test set, then calculate both the \textit{purity} (the mean intra-cluster similarity) and \textit{colocation} (the mean inter-cluster similarity) for the clusters of sentences assigned to each subpath $q_{1:d}$ for $d\leq4$. Finally, we take the difference between the purity and colocation as an overall measure of the quality of the representation space.

We compare to Hercules \citep{hosking-etal-2023-attributable} and  a baseline using recursive \mbox{$k$-means} over embeddings from a pretrained embeddings model \cite[MiniLM,][]{reimers-gurevych-2019-sentence}. We apply $k$-means to the embeddings, then calculate the residual errors between cluster centroids and embeddings, apply $k$-means to those errors, and repeat. All methods use the same number of clusters at each level ($K=12$).

\begin{figure}[t!]
    \centering
    \includegraphics[width=0.45\textwidth]{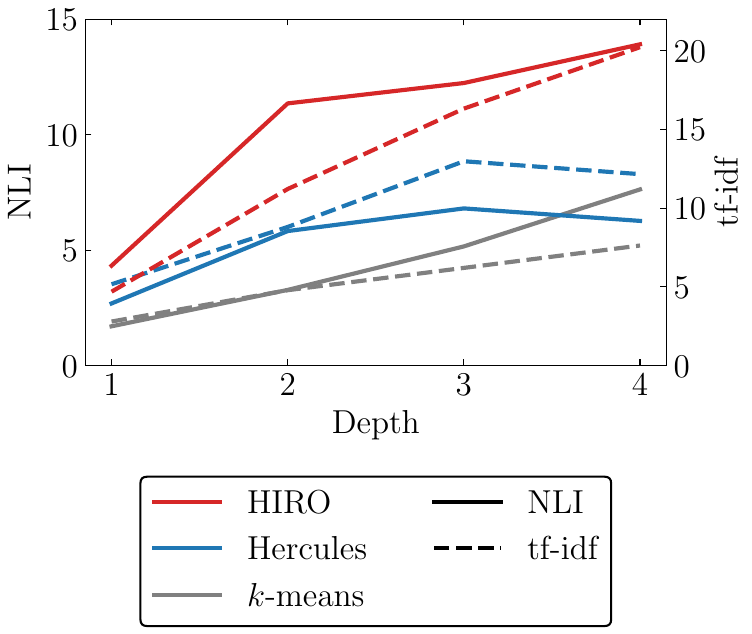}  
    \caption{Cluster quality by depth, as measured by the difference between cluster purity and colocation for the \textsc{Space} test set, according to NLI (solid line) and tf-idf (dashed line) similarity measures. HIRO learns a higher quality encoding space than comparison methods.}
    \vspace{-0.2cm}
    \label{fig:encoding_eval}
    \vspace{-0.2cm}
\end{figure}

\paragraph{Model Configuration} We use a 6 layer Transformer, with token embeddings initialized from BERT base \citep{devlin-etal-2019-bert}.\footnote{We experimented with using BERT as the encoder but found no significant improvement, since the discrete encoding is the main bottleneck in the model.} We set the codebook size $K = 12$, with the number of levels $D = 12$, based on validation set performance. Other hyperparameters are given in \Cref{appx:hparams}.

\paragraph{HIRO learns a higher quality encoding space}
\Cref{fig:encoding_eval} shows the overall quality (using both NLI and tf-idf measures of similarity), indicating that HIRO learns a more semantically distributed space at all levels of depth than comparison approaches, according to both similarity measures. The separation between clusters increases with depth, confirming that the encoder learns a semantic hierarchy. We believe these results indicate that our method could potentially be used for more general purpose document retrieval \cite[similar to][]{10.1145/3539618.3591651}, which is beyond the scope of this paper.

\section{Retrieving Popular Opinions}

Recall that a good summary should include opinions that occur repeatedly in the input reviews, as well as opinions that differentiate the current entity from others. 
In \Cref{sec:part1} we showed how the HIRO encoder maps a single sentence $\mathbf{x}$ to a path $q_{1:D}$. We now exploit the discrete, hierarchical nature of the representation space to index a large number of review sentences, then identify informative sentences to use to generate a summary. We hypothesize that our content selection method leads to clusters that better represent the balance of opinions in the input reviews.

\subsection{Method}

For each review $\mathbf{R} \in \mathcal{R}_e$ about an entity $e \in \mathcal{E}$, we separately encode each sentence within the review to its path $q_{1:D}$, giving an indexed review $\mathbf{Q}(\mathbf{R})$.

Our content selection method identifies the parts of the hierarchy that are particularly popular for each entity, and extracts the corresponding clusters of sentences. This process is query-free; instead, we assign each subpath in the hierarchy $q_{1:d}$ a score based on its popularity within the indexed input reviews, and `retrieve' all sentences that were mapped to the $k$ highest-scoring subpaths.
Our scoring function is inspired by \mbox{tf-idf} \citep{tfidf}, which is designed to measure the importance of a particular term with respect to a set of baseline documents. We define the \textit{term popularity} $\tp(q_{1:d}, e)$ of a path as the fraction of indexed reviews for entity $e$ which contain the subpath $q_{1:d}$, 
\begin{align}
    \tp(q_{1:d}, e) =& \frac{1}{ |\mathcal{R}_e| }\sum_{\textbf{R} \in \mathcal{R}_e} \mathbb{I}\big(q_{1:d} \in \mathbf{Q}(\textbf{R}) \big) ,
    \label{eq:tp}
\end{align}
where $\mathbb{I}$ is the indicator function.
We define the \textit{inverse baseline popularity} $\ibp$ as the reciprocal of the mean term popularity across \textit{all entities} $\mathcal{E}$, 
\begin{align}
    \ibp(q_{1:d}) =& \biggl( \frac{\alpha + \sum_{e \in \mathcal{E}} \tp(q_{1:d}, e)}{\alpha + |\mathcal{E}|} \biggr)^{-1} ,
    \label{eq:ibp}
\end{align}
where the smoothing constant $\alpha$ allows us to  balance between absolute and comparative popularity. The overall score is then
\begin{align}
    \text{score}(q_{1:d}, e) =& \tp(q_{1:d}, e) \times \ibp(q_{1:d}) .
    \label{eq:scoring}
\end{align} 
Intuitively, the score represents the relative popularity within the current entity of a path $q_{1:d}$ compared to all entities in the dataset.

Our scoring scheme conveniently accounts for the fact that short paths are more common;\footnote{The presence of a path $q_{1:D}$ for a review also implies the presence of all subpaths $q_{1:d}, d<D$.} short paths will also have a higher baseline popularity, leading to an overall score that is effectively normalized for the depth of the subpath.

After retrieving the clusters of sentences corresponding to the top-$k$ highest scoring subpaths, we filter out sentences with very low lexical similarity to other cluster members, and combine any clusters of sentences with high lexical overlap.


\subsection{Evaluation}
\label{sec:clustereval}

We evaluate whether our method successfully selects clusters of sentences containing informative opinions, using reviews from \textsc{Space}, as in \Cref{sec:part1_eval}. First, we measure the overlap between retrieved clusters and \textit{oracle clusters} constructed by selecting sentences with high overlap to the reference summaries,\footnote{\textsc{Space} includes multiple reference summaries for each entity; we randomly select one when determining the oracle clusters.} using the Adjusted Rand Index \cite[ARI,][]{ari}. Second, we evaluate the average \textit{prevalence} \citep{prevalence} of sentences in retrieved clusters, which uses a NLI model \cite[ALBERT, trained on VitC;][]{albert,schuster-etal-2021-get} to evaluate how many input reviews support each retrieved sentence. 

We compare HIRO to the clusters extracted by Hercules \citep{hosking-etal-2023-attributable}, and the oracle upper bound. As a baseline, we apply $k$-means to MiniLM embeddings \citep{reimers-gurevych-2019-sentence}, then extract the 25 sentences whose embeddings are closest to each centroid. For HIRO, we select the top $k=8$ subpaths for each entity, and set the smoothing constant $\alpha=6$. 

\paragraph{HIRO selects higher prevalence sentences}

\Cref{tab:cluster_eval} confirms that HIRO retrieves clusters that more closely match the oracle clusters, and contain opinions that are more prevalent in the input reviews compared to prior work. The oracle ARI score is less than 1 because some sentences appear multiple times in different clusters.

\begin{table}[t!]
    \centering
    \small
    \begin{tabular}{l|cc}
        \textbf{System} & \textbf{Prevalence} & \textbf{ARI} \\
        \hline\hline
        $k$-means & 38.1 & 0.59 \\
        Hercules & 32.3 & 0.50 \\
        HIRO & \textbf{46.5} & \textbf{0.69} \\
        \hline
        (Oracle) & 48.1 & 0.73 \\
    \end{tabular}
    \vspace{-0.2cm}
    \caption{Evaluation of our cluster selection method, compared to a range of baseline approaches. HIRO selects clusters of sentences that more closely match the references and contain more prevalent opinions.}
    \vspace{-0.2cm}
    \label{tab:cluster_eval}
\end{table}

\section{Generating Coherent Summaries}

Given the retrieved clusters of sentences for each entity, we want to generate a coherent and fluent textual summary. LLMs are well suited to this constrained rewriting task, and we leverage the zero-shot abilities of instruction-tuned models to map clusters of sentences to a readable summary.

\subsection{Method}

We generate a summary from the retrieved clusters in three ways, with varying trade-offs between coherence and attributability.
\paragraph{HIRO\textsubscript{ext}} We generate \textit{extractive} summaries by selecting the centroid of each cluster; we compute all pairwise ROUGE-2 scores between sentences in each cluster, and choose the sentence with highest average similarity to other cluster members. This approach is inherently attributable, since each summary sentence is extracted verbatim from a review.
\paragraph{HIRO\textsubscript{sent}} We generate summaries one sentence at a time by passing the contents of a single cluster as input to an instruction-tuned LLM with a simple prompt that requests a single sentence that summarizes the main points. This leads to more fluent output that is likely to be attributable, since each output sentence has an associated cluster of evidential sentences used to generate it.
\paragraph{HIRO\textsubscript{doc}} We generate summaries as a single document, by passing the sentences from all retrieved clusters for an entity to the LLM in one go. This gives summaries that are more coherent and less redundant, since the LLM has control over the whole summary. However, it is not possible to identify which cluster was used to generate each part of the summary, and therefore more difficult to determine the attributability of the output.

The ideal balance between coherence and the granularity of associated evidence is likely to vary by application.

\paragraph{Experimental Setup}

For the variants that require an LLM, we use Mistral 7B Instruct v0.2 to generate summaries from retrieved clusters. Mistral 7B was chosen based on its qualitative performance during model development, but we compare using alternative models in \Cref{sec:ablations}. The LLM is queried in a zero-shot manner, and the prompts used are given in \Cref{appx:prompts}. We sample with a temperature of $0.7$, and report the mean and standard deviation scores based on 3 samples.

\newcolumntype{H}{>{\setbox0=\hbox\bgroup}c<{\egroup}@{}}

\begin{table*}[t!]
    \centering
    \small
    \begin{tabular}{@{~}c@{~~}l@{~}||@{~}r@{~}r@{~}|@{~}r@{~}r@{~}r@{~}||@{~}r@{~}r@{~}|@{~}r@{~}r@{~}r@{~}}
     & & \multicolumn{5}{@{~}c@{~}||@{~}}{\textbf{\textsc{Space}}} & \multicolumn{5}{c}{\textbf{{AmaSum} (4 domains)}} \\
    & \textbf{System} & {R-2} $\uparrow$  & {{R-L}} $\uparrow$  & {Prev.} $\uparrow$ & {Gen.} $\downarrow$ & {SAP} $\uparrow$ &  {R-2} $\uparrow$ & {{R-L}} $\uparrow$ & {Prev.} $\uparrow$ & {Gen.} $\downarrow$ & {SAP} $\uparrow$ \\
\hline \hline

\multirow{7}{*}{\rotatebox{90}{\textit{Extractive}}}  

 & Rand. Review &  6.2\fs &  17.1\fs &  18.0\fs &  12.5\fs &  11.8\fs &  1.0\fs &  9.5\fs &  16.3\fs &  8.0\fs &  12.3\fs  \\ 
  & $k$-means &  9.5\fs &  19.8\fs &  27.9\fs &  25.0\fs &  15.4\fs &  2.3\fs &  12.0\fs &  14.9\fs &  11.4\fs &  9.2\fs  \\ 
 & LexRank &  5.9\fs &  16.4\fs &  18.2\fs &  \textbf{4.4}\fs &  16.0\fs &  2.7\fs &  12.2\fs &  9.0\fs &  \textbf{3.0}\fs &  7.5\fs  \\ 
 & QT &  10.3\fs &  21.5\fs &  24.9\fs &  23.3\fs &  13.3\fs &  1.5\fs &  11.4\fs &  10.9\fs &  7.3\fs &  7.3\fs  \\ 
 & SemAE &  11.1\fs &  23.5\fs &  29.2\fs &  17.1\fs &  20.6\fs &  1.6\fs &  11.3\fs &  8.7\fs &  4.1\fs &  6.7\fs  \\ 
 & Hercules\textsubscript{ext} &  \textbf{13.2}\fs &  \textbf{24.4}\fs &  30.2\fs &  25.2\fs &  17.6\fs &  \textbf{3.0}\fs &  12.5\fs &  9.5\fs &  6.7\fs &  6.2\fs  \\ 
 & HIRO\textsubscript{ext} &  11.7\fs &  22.1\fs &  \textbf{36.3}\fs &  20.5\fs &  \textbf{26.1}\fs &  2.7\fs &  \textbf{12.6}\fs &  \textbf{19.4}\fs &  9.5\fs &  \textbf{14.6}\fs  \\ 

 \hline
 \multirow{9}{*}{\rotatebox{90}{\textit{Abstractive}}}  
  & CopyCat &  12.1\fs &  22.9\fs &  \textbf{48.3}\fs &  70.9\fs &  12.9\fs &  1.5\fs &  11.2\fs &  15.8\fs &  21.0\fs &  5.3\fs  \\ 

 & BiMeanVAE &  13.7\fs &  27.1\fs &  45.0\fs &  61.4\fs &  14.2\fs &  2.0\fs &  12.5\fs &  14.7\fs &  24.1\fs &  2.7\fs  \\ 
 & COOP &  14.2\fs &  \textbf{27.2}\fs &  46.1\fs &  63.2\fs &  14.5\fs &  2.8\fs &  14.1\fs &  \textbf{18.8}\fs &  30.3\fs &  3.7\fs  \\ 
 & Hercules\textsubscript{abs} &  \textbf{14.8}\fs &  \textbf{27.2}\fs &  32.2\fs &  36.1\fs &  14.1\fs &  2.0\fs &  11.8\fs &  8.5\fs &  9.2\fs &  3.9\fs  \\ 
 & Zero-shot Mistral 7B &  5.3{\tiny$\pm$0.1}  &  19.6{\tiny$\pm$0.4}  &  41.3{\tiny$\pm$1.3}  &  34.3{\tiny$\pm$3.3}  &  24.2{\tiny$\pm$0.8}  &  1.9{\tiny$\pm$0.0}  &  12.6{\tiny$\pm$0.0}  &  17.3{\tiny$\pm$0.2}  &  17.6{\tiny$\pm$0.4}  &  8.5{\tiny$\pm$0.2}   \\ 
 
  & HIRO\textsubscript{sent}+Mistral 7B &  4.5{\tiny$\pm$0.1}  &  18.2{\tiny$\pm$0.0}  &  36.2{\tiny$\pm$0.8}  &  \textbf{20.1}{\tiny$\pm$0.5}  &  26.2{\tiny$\pm$0.9}  &  3.5{\tiny$\pm$0.0}  &  14.1{\tiny$\pm$0.1}  &  14.6{\tiny$\pm$0.3}  &  \textbf{6.9}{\tiny$\pm$0.1}  &  \textbf{11.2}{\tiny$\pm$0.3}   \\ 
 & HIRO\textsubscript{doc}+Mistral 7B &  7.0{\tiny$\pm$0.2}  &  20.5{\tiny$\pm$0.3}  &  44.0{\tiny$\pm$3.0}  &  28.8{\tiny$\pm$2.1}  &  \textbf{29.6}{\tiny$\pm$2.1}  &  \textbf{4.0}{\tiny$\pm$0.0}  &  \textbf{15.1}{\tiny$\pm$0.1}  &  15.3{\tiny$\pm$0.1}  &  9.4{\tiny$\pm$0.3}  &  10.6{\tiny$\pm$0.1}   \\
 \hline
 & (References) &  ---\fs &  ---\fs &  44.3\fs &  50.2\fs &  19.2\fs &  ---\fs &  ---\fs &  9.3\fs &  7.0\fs &  5.8\fs  \\ 
 & (Oracle) &  45.0\fs &  53.3\fs &  41.0\fs &  38.5\fs &  21.7\fs &  14.4\fs &  26.0\fs &  12.3\fs &  9.0\fs &  7.8\fs  \\
    \hline\hline

    \end{tabular}
    \caption{Results for automatic evaluation of summary generation on the test splits. R-2 and R-L represent ROUGE-2/L F1 scores. Prev. refers to Prevalence, Gen. refers to Genericness, and SAP refers to Specificity-Adjusted Prevalence. ROUGE scores are no longer considered reliable \citep{callison-burch-etal-2006-evaluating,tay-etal-2019-red,10.1162/tacl_a_00373}, so we consider SAP to be our primary metric. The best scores for extractive and abstractive systems are shown in bold. Results for systems involving LLMs are based on 3 samples, with the mean and standard deviation shown. HIRO generates summaries with the best balance between prevalent opinions and specificity.}
     \vspace*{-.2cm}
    \label{tab:automatic_general}
\end{table*}

\subsection{Automatic Evaluation}
\label{sec:part3_autoeval}

Human evaluation is the gold standard (\Cref{sec:part3_humeval}), but automatic metrics remain useful for model development. ROUGE scores are no longer reliable (\Cref{sec:relatedwork}), but we nonetheless report them for consistency with prior work. \citet{prevalence} propose a \textit{prevalence} metric, that uses an NLI model to determine how many input reviews contain supporting evidence for each sentence in the summary, but this suffers from a failure mode that overly rewards generic statements. A good summary should include common opinions, but should also help a user to differentiate between multiple entities.


To counteract this pathology, we propose a modified version of prevalence, that explicitly penalizes generic summaries. First, we define the \textit{genericness} of a summary as the average number of summaries from \textit{other entities} that support each sentence in a given summary, as measured by an NLI model \cite[ALBERT, trained on VitC;][]{albert,schuster-etal-2021-get}. Then, we define the Specificity Adjusted Prevalence score (SAP) as 
\begin{equation}
\texttt{SAP} = \texttt{prevalence} - \alpha~\texttt{genericness},    
\end{equation}
where $\alpha$ is a constant that controls the balance between absolute prevalence and specificity. In practice, the ideal summary is unlikely to be entirely unique and a user may want to allow some degree of overlap between generated summaries. We report the prevalence and genericness, as well as the combined SAP with $\alpha=0.5$.

\paragraph{Datasets} We evaluate summary generation using \textsc{Space} (\Cref{sec:part1_eval}), which includes multiple reference summaries created by human annotators for each entity. We also include \textbf{AmaSum} \citep{brazinskas-etal-2021-learning}, to evaluate summary generation on a wide range of categories of Amazon products, with an average of 560 reviews per entity. The reference summaries were collected from professional review websites, and therefore are not necessarily grounded in the input reviews. We use the same splits, based on four product categories, as released by \citet{hosking-etal-2023-attributable}. Further dataset statistics are given in \Cref{appx:datasets}.

\paragraph{Comparison Systems}
We select a \textbf{random review} from the inputs as a lower bound. 
We include an extractive \textbf{oracle} as an upper bound, by selecting the input sentence with highest ROUGE-2 similarity to each reference sentence.\footnote{When multiple references are available, we select one at random.} For a \textbf{k-means} baseline, we run $k$-means on MiniLM sentence embeddings \citep{reimers-gurevych-2019-sentence}, then extract the nearest sentence to the cluster centroids. We set $k=8$ to match the average sentence length of the reference summaries.
\textbf{Lexrank} \citep{lexrank} is an unsupervised extractive method using graph-based centrality scoring of sentences.
\textbf{QT} \citep{angelidis-etal-2021-extractive} uses vector quantization to map sentences to a discrete encoding space, then generates extractive summaries by selecting representative sentences from clusters.
\textbf{SemAE} \citep{basu-roy-chowdhury-etal-2022-unsupervised} is an extractive method that extends QT, relaxing the discretization and encoding sentences as mixtures of learned embeddings.
\textbf{CopyCat} \citep{brazinskas-etal-2020-unsupervised} is an abstractive approach that models sentences as observations of latent variables representing entity opinions, trained in a `leave one out' manner.
\textbf{BiMeanVAE} and \textbf{COOP} \citep{iso-etal-2021-convex-aggregation} are abstractive methods that encode full reviews as continuous latent vectors using an autoencoder, and take the average (BiMeanVAE) or an optimised combination (COOP) of review encodings.
We compare to a recent open weight instruction-tuned \textbf{LLM}, specifically Mistral Instruct v0.2 7B \citep{jiang2023mistral}. Since no training data is available, the LLM was prompted zero-shot as per \Cref{appx:prompts}, and sampled with temperature $0.7$. We report the mean and standard deviation scores based on 3 samples.

Most of the abstractive methods are not scalable and have upper limits on the number of input reviews. CopyCat and Mistral 7B have a maximum input sequence length, while COOP exhaustively searches over combinations of input reviews. We use 8 randomly selected reviews as input to CopyCat, COOP, and Mistral 7B.

\paragraph{HIRO gives the best balance between prevalence and specificity}

The results in \Cref{tab:automatic_general} show that HIRO achieves the highest SAP scores across both datasets, indicating that it generates summaries with the best balance between absolute prevalence and specificity.  While CopyCat and COOP achieve the highest prevalence scores on \textsc{Space} and AmaSum respectively, they also display some of the highest genericness; qualitatively, the outputs are very similar to each other, with few specific details. We give example output in Tables~\ref{tab:output} and \ref{tab:output2}.


\paragraph{References are not the upper bound} 

While the oracle summaries score highly in terms of specificity-adjusted prevalence, some systems (including HIRO) outperform them. This indicates the difficulty with constructing reference summaries for entities with large numbers of reviews; it is not feasible for human annotators to reliably summarize such a large amount of information.

\begin{table}[!t]
\small
    \centering
    \begin{tabular}{l|cc}
    \textbf{System} & \textbf{\% Partial} & \textbf{\% Majority} \\
    \hline
       Hercules\textsubscript{abs}  & 85.4 & 27.6 \\
        HIRO\textsubscript{sent} & 81.6 & 21.6 \\
        HIRO\textsubscript{doc} &  \textbf{91.8} & \textbf{28.4}
    \end{tabular}
    \caption{Results for automatic evaluation of the evidence supplied by attributable systems, showing the percentage of summary sentences that have support from at least one sentence in the evidence set (partial support) and from at least half the sentences in the evidence (majority support). HIRO generates summaries that have strong partial support from the associated evidence sets, with improved majority support compared to Hercules.} 
    \label{tab:attribution}
\end{table}

\begin{figure}[!t]
    \centering
    \includegraphics[width=0.4\textwidth]{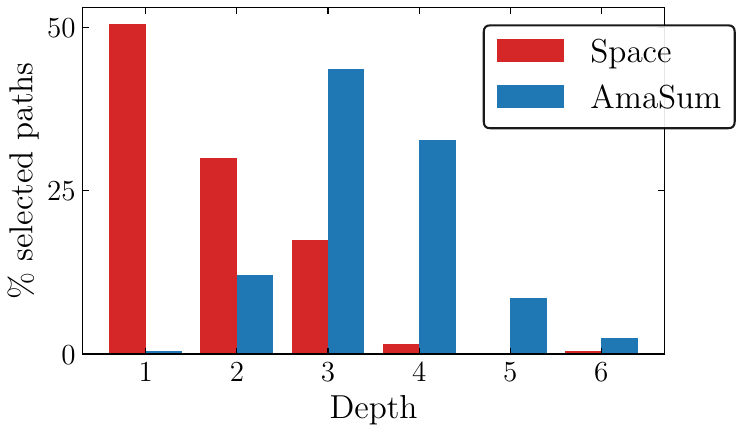}  
    \vspace{-0.2cm}
    \caption{Distribution of selected subpaths by depth. A large fraction of the extracted clusters come from paths deeper than the top level.}
    \vspace{-0.2cm}
    \label{fig:path_dist}
    
\end{figure}

\begin{figure*}[t!]
    \centering
    \includegraphics[width=0.98\textwidth]{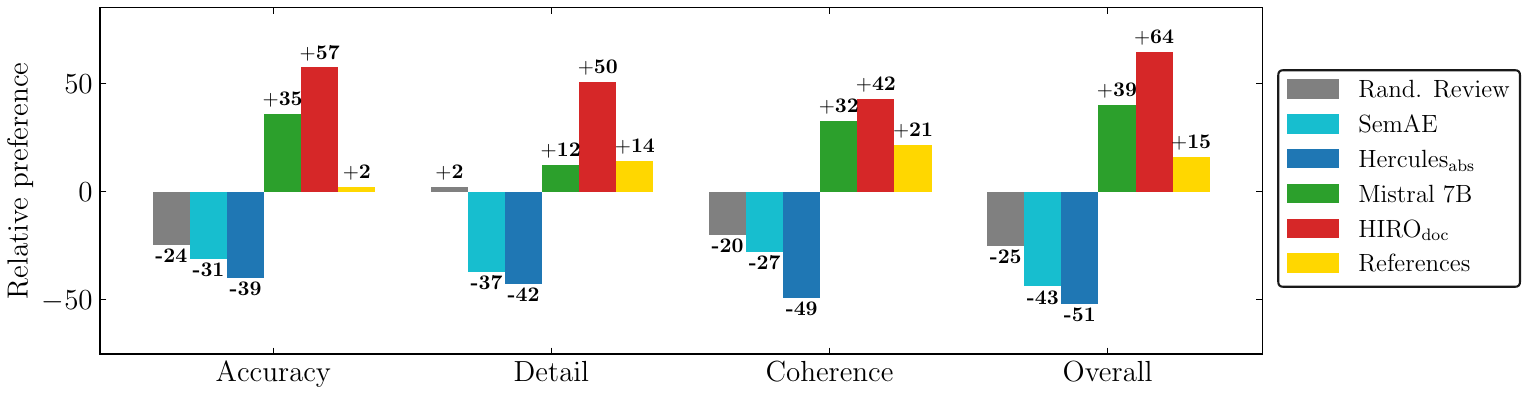}  

    \vspace{-.2cm}
    \caption{Overall results of our human evaluation based on the test splits of \textsc{Space} and AmaSum. Crowdworkers were asked for pairwise preferences between generated summaries in terms of their Accuracy, Detail, and Coherence \& Fluency, as well as an overall preference. HIRO generates more coherent and detailed summaries that better represent the opinions in the input reviews than comparison systems, and are preferred by human annotators.}
    \vspace{-.2cm}
    \label{fig:human_eval}
\end{figure*}

\paragraph{HIRO is more faithful to selected evidence} 

To evaluate how faithful the generated summaries are to the retrieved sentence clusters or \textit{evidence sets}, we use an NLI model to determine how many sentences in each cluster either entail or are entailed by the corresponding sentence in the output summary, and take the mean. Considering both forward and backward entailment in this manner accounts for the different levels of granularity between the inputs and summary \citep{zhang-etal-2024-fine}; input reviews are likely to be more specific than summary sentences, but concise summary sentences are likely to contain multiple assertions, e.g. ``The food was good \textit{and} the rooms were clean''. HIRO\textsubscript{doc} does not align the evidence sets with each generated sentence, so we calculate the maximum support score over all sets for each summary sentence. Most abstractive systems are not attributable, and so we only compare with Hercules. \Cref{tab:attribution} shows the percentage of summary sentences that have support from at least \textit{one} sentence in the evidence (partial support) and from at least \textit{half} the sentences in the evidence (majority support). HIRO\textsubscript{doc} generates summaries that have strong partial support from the associated evidence sets, with improved majority support compared to Hercules despite also being significantly more detailed. 

\paragraph{HIRO makes use of the hierarchy} We confirm that HIRO exploits the hierarchical nature of the representation space. \Cref{fig:path_dist} shows the distribution of selected subpath depths for both datasets, indicating that HIRO takes advantage of the hierarchy and selects clusters deeper than the top level. This is particularly true for AmaSum, where there is a wider range of product types, causing the selection process to search deeper in the tree.

\subsection{Human Evaluation}
\label{sec:part3_humeval}

We conduct a human evaluation to verify that HIRO generates summaries that are coherent and accurately reflect the opinions in the input reviews. We recruit crowdworkers through Prolific, selected to be L1 English speakers from the US or UK with a minimum of 100 previously approved studies, compensated above the UK living wage at 12GBP/hr. Participants were allowed to rate at most 5 samples. We show annotators a set of 50 reviews (chosen based on pilot studies to balance annotator load with reliability), followed by two generated summaries. We solicit pairwise preferences \citep{louviere1990best} along three dimensions, as well as an overall preference: 
\begin{itemize}
\itemindent=-8pt

\vspace{-0.4em}
\setlength\itemsep{-0.3em}
    \item \textbf{Accuracy} ---  Which summary accurately reflects the balance of opinion in the  reviews?
    \item \textbf{Detail} --- Which summary includes more specific details?
    \item \textbf{Coherence} --- Which summary is easy to read and avoids contradictions?
    \item \textbf{Overall} --- Which summary do you think is better, overall?
    \vspace{-0.4em}
\end{itemize}

The full instructions are reproduced in \Cref{appx:annotation}. Ties (i.e., `no difference') were allowed. We gather annotations for 10~entities each from the \textsc{Space} and AmaSum test sets, with 3~annotations for each pairwise combination of system outputs, leading to a total of 900~pairwise ratings. The study was approved by ethics committee, ref. \#491139.

We compare HIRO\textsubscript{doc} to the top performing extractive and abstractive systems from \Cref{tab:automatic_general}, SemAE and Hercules\textsubscript{abs}. HIRO\textsubscript{doc} uses Mistral 7B as the generator, so we also compare to Mistral 7B \textit{without} HIRO (i.e., prompted directly with reviews). Finally, we include a random review as a lower bound, and the references as an upper bound.

\paragraph{Humans prefer summaries generated by HIRO}

The results in \Cref{fig:human_eval} show that HIRO\textsubscript{doc} produces summaries that outperform comparison systems across all dimensions, producing summaries that coherently and accurately represent the opinions in the input reviews. Differences between HIRO\textsubscript{doc} and other systems are significant in all cases (using a one-way
ANOVA with post-hoc Tukey HSD test, $p<0.05$), except for coherence versus Mistral 7B. Both Mistral 7B and HIRO\textsubscript{doc} outperform the reference summaries, supporting findings from prior work \citep{bhaskar-etal-2023-prompted,hosking2023human}.

\begin{figure}[t!]
    \centering
    \includegraphics[width=0.45\textwidth]{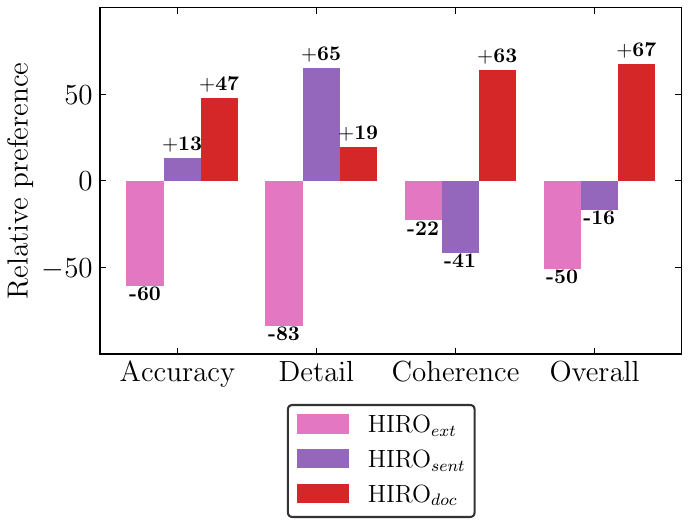}  

    \caption{Results of a human evaluation comparing the three variants of HIRO: extractive, sentence-wise and document. Overall, annotators prefer the coherence of the document approach, but the sentence-wise variant generates more detailed summaries that are also more attributable. The preferred tradeoff between coherence and attribution will vary depending on the application.}
    \label{fig:human_eval_hiro}
\end{figure}

\begin{table*}[t!]
    \centering
    \small
    \begin{tabular}{@{~}l@{~}l@{~~}l@{~}||@{~}r@{~}r@{~}|@{~}r@{~}r@{~}r@{~}||@{~}r@{~}r@{~}|@{~}r@{~}r@{~}r@{~}}
    & & & \multicolumn{5}{@{~}c@{~}||@{~}}{\textbf{\textsc{Space}}} & \multicolumn{5}{c}{\textbf{{AmaSum}}} \\
     & \textbf{Clusters} & \textbf{LLM} & R-2 $\uparrow$ & R-L $\uparrow$ &  {Prev.} $\uparrow$ & {Gen.} $\downarrow$ & {SAP} $\uparrow$ & R-2 $\uparrow$ & R-L $\uparrow$ &   {Prev.} $\uparrow$ & {Gen.} $\downarrow$ & {SAP} $\uparrow$ \\
\hline \hline
 \multirow{3}{*}{\textit{Llama 2 7B}} & Zero-shot & Llama 2 7B &  4.4 &  17.6 &  32.6 &  25.7 &  19.7 &  1.5 &  11.5 &  \textbf{17.7} &  17.6 &  8.9  \\ 
 & HIRO\textsubscript{sent} &  Llama 2 7B &  5.2 &  16.9 &  36.2 &  \textbf{20.0} &  26.2 &  3.6 &  14.0 &  14.6 &  \textbf{6.3} &  \textbf{11.5}  \\ 
 & HIRO\textsubscript{doc} &  Llama 2 7B &  \textbf{6.9} &  \textbf{19.9} & \textbf{ 48.5} &  29.9 &  \textbf{33.5} &  \textbf{4.0} &  \textbf{15.2} &  16.0 &  12.1 &  10.0  \\ 
 \hline
 \multirow{3}{*}{\textit{Llama 2 13B}} & Zero-shot & Llama 2 13B &  6.3 &  19.1 &  36.7 &  28.1 &  22.6 &  2.3 &  13.1 &  18.1 &  14.3 &  10.9  \\ 
 
 &  HIRO\textsubscript{sent} &  Llama 2 13B &  6.6 &  18.8 &  35.0 &  \textbf{23.7} &  23.1 &  3.8 &  14.3 &  13.5 &  \textbf{6.2} &  10.4  \\ 
 & HIRO\textsubscript{doc} &  Llama 2 13B &  \textbf{8.5} &  \textbf{21.7} &  \textbf{46.2} &  27.0 &  \textbf{32.7} &  \textbf{4.3} & \textbf{ 15.7} &  \textbf{19.0} &  9.9 &  \textbf{14.1}  \\ 
 \hline
 \multirow{3}{*}{\textit{Llama 2 70B}} &Zero-shot & Llama 2 70B &  5.6 &  19.4 &  \textbf{48.9} &  37.2 &  30.3 &  2.1 &  12.5 &  16.8 &  22.4 &  5.6  \\ 
 & HIRO\textsubscript{sent} &  Llama 2 70B &  5.8 &  18.5 &  41.1 &  \textbf{19.6} &  31.3 &  3.6 &  14.5 &  15.1 &  \textbf{3.7} &  \textbf{13.2}  \\ 
 & HIRO\textsubscript{doc} &  Llama 2 70B & \textbf{ 8.3} &  \textbf{21.3} &  48.5 &  30.0 &  \textbf{33.5} &  \textbf{4.4} &  \textbf{16.0} &  \textbf{17.8} &  9.9 &  12.9  \\ 
\hline\hline
 \multirow{3}{*}{\textit{Sent-wise}} & $k$-means & Mistral 7B &  \textbf{4.5} &  17.1 &  30.1 &  30.8 &  14.7 &  3.2 &  13.3 &  12.9 &  12.4 &  6.7  \\ 
 & Hercules &  Mistral 7B &  3.9 &  17.0 &  27.4 &  25.8 &  14.5 &  3.3 &  11.8 &  8.5 &  8.7 &  4.1  \\ 
 & HIRO\textsubscript{sent} &  Mistral 7B &  \textbf{4.5} &  \textbf{18.2} &  \textbf{36.4} &  \textbf{20.2} &  \textbf{26.3} &  \textbf{3.5} &  \textbf{14.1} &  \textbf{14.6} &  \textbf{6.9} &  \textbf{11.2}  \\ 
 \hline
 \multirow{3}{*}{\textit{Doc-wise}} & $k$-means & Mistral 7B &  6.4 &  20.5 &  40.2 &  36.3 &  22.0 &  3.7 &  14.5 &  \textbf{16.2} &  13.9 &  9.2  \\ 
 & Hercules & Mistral 7B &  \textbf{7.6} & \textbf{ 21.0} &  42.9 &  39.2 &  23.3 &  3.8 &  14.7 &  13.8 &  11.6 &  8.0  \\ 
 & HIRO\textsubscript{doc} &  Mistral 7B &  7.0 &  20.5 &  \textbf{44.0} &  \textbf{28.8} &  \textbf{29.6} &  \textbf{4.0} &  \textbf{15.1} &  15.3 &  \textbf{9.4} & \textbf{ 10.6 } \\

 \hline\hline

    \end{tabular}
    
    \vspace{-.1cm}
    \caption{Automatic evaluations comparing HIRO to zero-shot summarization, using a range of different LLMs, for the test splits. We show the mean scores based on 3 samples, and best scores within each comparison are bolded. In all cases, HIRO improves on the zero-shot approach. We also compare HIRO to other cluster selection methods, finding that HIRO leads to summaries with a better balance between prevalance and specificity.}
    \vspace{-.1cm}
    \label{tab:hiro:ablations}
\end{table*}

\paragraph{Comparing HIRO variants} We run an additional human evaluation to evaluate the tradeoff between quality and attribution between the three HIRO variants, collecting annotations for 5 entities from each dataset from 3 annotators each. \Cref{fig:human_eval_hiro} shows that annotators prefer the output of HIRO\textsubscript{doc}, although the sentence-wise variant generates more detailed summaries. The preferred tradeoff between attribution and coherence will likely vary depending on the application.

\begin{table*}[ht!]
\renewcommand{\arraystretch}{1.2}
    \centering
    \small
    \begin{tabular}{@{~}m{2cm}@{~}|@{~}m{13.5cm}@{~}}
    \textbf{System} & \textbf{Output} \\
    \hline\hline
\textit{Reference} & 
The staff were very friendly, courteous and helpful. The hotel and room was clean. The rooms are very spacious with a nice oceanfront view and a gigantic balcony! Pool bar is great and food is good, too. The hotel has a perfect location. It was very nicely located with a 24 hours cafe downstairs, 2 pools , 2 bars and some beach stores to shop in.
 \\ 
 \hline  
\textit{Mistral 7B} & 
The Grand Hotel in Ocean City, Maryland, consistently receives positive reviews for its exceptional customer service, cleanliness, and convenient location on the boardwalk. Rooms offer great views, and amenities include a spa, restaurant, lounge, indoor pool, game room, and exercise room. Guests appreciate the family-friendly atmosphere and the ease of having everything they need within the hotel. \sethlcolor{Orange!30}\hl{Some reviewers praise Nina Railean} for her helpfulness and accommodating nature. However, \sethlcolor{LimeGreen!30}\hl{one review} warns against staying during Memorial Day weekend due to loud parties.
 \\ 
 \hline  
\textit{HIRO\textsubscript{sent}} & 
All rooms offer excellent ocean views at this hotel. The Grand Hotel \& Spa offers direct ocean views and easy beach access. The Grand Hotel \& Spa offers clean and spacious rooms. No problems with accommodations or access despite rain and floor issues. The Grand Hotel \& Spa offers beautiful views and a comfortable stay with excellent service. The Grand Hotel \& Spa exceeded expectations. Rooms need updating peeling wallpaper, large, dark interiors.
 \\ 
 \hline  
\textit{HIRO\textsubscript{doc}} & 
The Grand Hotel \& Spa offers stunning ocean views from most rooms, with some having a direct view of the boardwalk and beach. The hotel's design ensures that guests can enjoy the ocean despite room location. The beach is clean, and the location is convenient for easy access. Rooms are described as clean and large, but some note a need for improvement in terms of lighting and room condition. Overall, guests had positive experiences and appreciated the ocean access.
 \\ 
 
    \end{tabular}
    \vspace*{-.1cm}
    \caption{Generated summaries for The Grand Hotel Maryland, from \textsc{Space}, comparing HIRO to zero-shot Mistral 7B. While all generated summaries are fluent and convincing, Mistral 7B makes reference to the opinion of a \sethlcolor{LimeGreen!30}\hl{single user}, and positive sentiment about \sethlcolor{Orange!30}\hl{a member of staff} that is not supported by the full set of reviews. These extrapolations highlight the problem with models that can only accept a limited number of reviews as input.}
    \label{tab:hiro:error}
\end{table*}

\begin{table*}[ht!]
\renewcommand{\arraystretch}{1.2}
    \centering
    \small
    \begin{tabular}{m{0.95\textwidth}}
        \hline\hline
         HIRO provides more specific detail about user frustrations and the product itself. Mistral 7B is quite generic. \\
         \hline
         Mistral 7B mentions that desk staff were unfriendly, but this is not substantiated by the reviews, the majority of which are overwhelmingly positive. It's also a contradiction since earlier 
         it says the staff were friendly.  \\
         \hline
         Only HIRO references the high failure rate. \\
         \hline
         HIRO seems more accurate in its appraisal of the staff; the reviews were mixed. However Mistral 7B is slightly more informative, particularly in relation to location and proximity to amenities. I feel like HIRO sits on the fence quite a lot, whereas Mistral 7B attempts to summarise better. \\
         \hline
         I think HIRO is the better written and reflects a broader view on the reviews but References isn't bad it does cover PS4 and gaming which alot of reviews were using it for. \\
         \hline
         SemAE is incoherent.
         HIRO does an excellent job of summarising everything, balancing all perspectives. \\
         \hline
         Can tell HIRO is AI generated \\
         \hline
         HIRO is by far the better written and has a little more detail and reflects the reviews more than Hercules\textsubscript{abs} \\
         \hline\hline
    \end{tabular}
    \caption{Comments from annotators for all pairs involving HIRO. Anonymised system labels have been replaced with the system names. The majority of comments are positive towards HIRO, although annotators comment that HIRO summaries may be less natural, or too conservative.}
    \label{tab:humancomments}
\end{table*}

\subsection{Analysis}

\paragraph{Ablations}
\label{sec:ablations}

\Cref{tab:hiro:ablations} compares HIRO to zero-shot prompting, for a range of different LLM sizes from the Llama 2 Chat family of LLMs \citep{touvron2023llama}. The same prompt template and hyperparameters were used as for Mistral 7B.  Increased parameter count does lead to improved SAP scores for the zero-shot approach on \textsc{Space}, but the largest 70B model scores worse on AmaSum than the 7B and 13B models.
For all choices of LLM, HIRO leads to summaries with a better balance between prevalence and specificity than zero-shot prompting; HIRO is likely to lead to improvements for any choice of instruction-tuned LLM.

We also compare using HIRO to identify clusters against a $k$-means baseline, and Hercules \citep{hosking-etal-2023-attributable}. For both datasets, using clusters selected with HIRO leads to summaries that are much less generic, while remaining comparatively prevalent. This confirms the results in \Cref{sec:clustereval}, indicating that HIRO selects more informative clusters of sentences than the comparison methods.

\paragraph{Qualitative Analysis}

\Cref{tab:hiro:error} shows output from Mistral 7B and HIRO as well as a reference summary, for The Grand Hotel Maryland from \textsc{Space}. While all generated summaries are fluent and convincing, Mistral 7B makes reference to the opinion of a \sethlcolor{LimeGreen!30}\hl{single user}, which should not appear in a summary. It also describes positive praise about \sethlcolor{Orange!30}\hl{a member of staff}, but a manual analysis shows that only 3 out of 5 reviews mentioning that individual are positive; the true sentiment is much more mixed than the summary indicates. These extrapolations highlight the problem with models that can only accept a limited number of reviews as input.

The examples of selected subpaths, sentence clusters and corresponding HIRO outputs in \Cref{tab:evidence_example} demonstrate the difficulty of evaluating attribution. The final cluster contains sentences that are all topically related, indicating that HIRO has learned a successful clustering. While the sentences and corresponding HIRO\textsubscript{sent} output are all broadly negative, it is not straightforward to determine whether the sentence ``Only one mirror in the room" counts as direct evidence towards the output ``Rooms were small, loud, and in need of renovation [...]". This partly explains the relatively low majority support scores in \Cref{tab:attribution}; some of the selected evidence may be consistent in topic and sentiment, but not directly entail the resulting output.

As well as collecting pairwise preferences in our human evaluation, we allowed annotators to leave qualitative comments. \Cref{tab:humancomments} shows all non-trivial
comments from annotators for pairs including HIRO, with anonymised labels replaced by the true model names. The majority of comments are positive towards HIRO, highlighting improved levels of detail and a better balance of the input reviews. However, some annotators note that HIRO summaries may be less natural or too conservative.


\section{Conclusion}

We propose HIRO, a modular method for unsupervised opinion summarization that uses a hierarchical index over sentences to select clusters of prevalent opinions. Our approach leverages pretrained Large Language Models to generate coherent and fluent summaries that are attributable and accurately reflect the popularity of opinions in the input reviews. Extensive experiments show that, as well as generating higher quality summaries, HIRO learns a more semantically distributed representation than competitive baselines. While we limit our experiments to opinion summarization, we believe that HIRO could be usefully applied to a wide range of other retrieval-augmented generation tasks.


\section*{Acknowledgements}

We thank the action editor and anonymous reviewers for their constructive feedback. This work was supported in part by the UKRI Centre for Doctoral Training in Natural Language Processing, funded by the UKRI (grant EP/S022481/1) and the University of Edinburgh. Lapata acknowledges the
support of the UK Engineering and Physical Sciences Research Council (grant EP/W002876/1).






\bibliography{anthology,custom}
\bibliographystyle{acl_natbib}

\appendix

\section{Annotation Instructions}
\label{appx:annotation}

\begin{quote}
    \textbf{Instructions}
    
In this task you will be presented with some reviews of a product/hotel, followed by two summaries produced by different automatic systems. Your task is to rate the system summaries based on the criteria listed below.

First, please skim read through the reviews, to try to get an overall idea of what opinions are mentioned frequently. Then, read the system summaries carefully, and rate each one according to the criteria.

Please read the criteria descriptions and system summaries carefully, and whenever is necessary re-read the summaries or reviews. You might want to use your browser's search function to help find parts of reviews that are relevant.

\textbf{Criteria}

\textit{Accuracy} -- Which system summary accurately reflects the balance of opinion in the input reviews? Statements in the summary should be backed up by multiple reviews.

\textit{Detail} -- Which system summary includes more specific details?


\textit{Coherence \& Fluency} -- Which system summary is easy to read and avoids contradictions?

\textit{Overall} -- Which system summary do you think is better, overall?
\end{quote}

\section{Hyperparameters}
\label{appx:hparams}

\begin{table}[ht]
    \centering
\small
    \begin{tabular}{@{~}l@{~}|@{~}l@{~}}
    \textbf{Parameter} & \textbf{Value} \\
    \hline\hline
    Embedding dim. $\mathbb{D}$ & 768 \\
    Encoder layers & 5 \\
    Feedforward dim. & 2048 \\
    Transformer heads & 8 \\
    Depth $D$& 12 \\
    Codebook size $K$& 12 \\
    Optimizer & Adam~\citep{adam} \\
    Learning rate & 1e-4 \\
    Batch size & 384 \\
    $\omega$ & 150 \\
    $\alpha_{init}$ & 0.5 \\
    $\tau_0$ & 1.0 \\
    $\tau_{min}$ & 0.5 \\
    $\gamma_{temp}$ & 33333 \\
    $\beta_{KL}$ & 0.0025 \\
    $\beta_{NL}$ & 0.05 \\
    $\gamma_{NL}$ & 1.5 \\
    top-$k$ subpaths & 8 \\
    $\tp-\ibp$ smoothing $\alpha$ & 6 (\textsc{Space}), 3 (AmaSum) 
    \end{tabular}
    \caption{Hyperparameter values used for our experiments.}
    \label{tab:hyperparams}
\end{table}


\section{LLM prompts}
\label{appx:prompts}

\renewcommand{\lstlistingname}{Prompt}

\begin{lstlisting}[caption=Baseline LLMs (Mistral 7B and all Llama 2 models),captionpos=b]
Review:
[...] (x8)
Write a summary in 70 words or less:\end{lstlisting}

\begin{lstlisting}[caption=HIRO\textsubscript{sent},captionpos=b]
Here is a list of sentences taken from reviews of the {entity name}:

[...]

In no more than 10 words, write a single concise sentence that includes the main point:
\end{lstlisting}

\begin{lstlisting}[caption=HIRO\textsubscript{doc},captionpos=b]
Here is a list of sentences taken from reviews of the {entity name}:

[...]

In no more than 60 words, write a concise summary that includes the main points:\end{lstlisting}

\section{Dataset Statistics}
\label{appx:datasets}


\begin{table}[!h]
    \centering
    \small
    \begin{tabular}{r|rr}
        & \textsc{Space} & AmaSum \\
        \hline\hline
        Entities & 8350 & 7255  \\
        Reviews & 303,357 & 533,972 \\
        Training pairs $(\mathbf{x},\mathbf{x}_{+})$ & 1,373,079 & 2,991,478  \\

    \end{tabular}
    \caption{Statistics for the training datasets. }
    \label{tab:dataset_stats_training}
\end{table}

\begin{table}[ht]
    \centering
    \small
    \begin{tabular}{r|rr}
        & \textsc{Space} & AmaSum \\
        \hline\hline
        Entities & 25 & 200  \\
        Reviews per entity & 100 & 560.4 \\
        Review length (words) & 162.6 & 49.9 \\
        Ref. summaries (words) & 82.0 & 80.1 \\
    \end{tabular}
    \caption{Statistics for the evaluation datasets.}
    \label{tab:dataset_stats_eval}
\end{table}

\section{Example Outputs}
\label{appx:examples}

See \Cref{tab:output}, \Cref{tab:output2} and \Cref{tab:evidence_example}.

\begin{table*}[ht]
\renewcommand{\arraystretch}{1.2}
    \centering
    \small
    \begin{tabular}{@{}m{2cm}@{~}|@{~}m{13.2cm}@{}}
    \textbf{System} & \textbf{Output} \\
    \hline\hline
\textit{Reference} & 
The staff were very welcoming and were always happy to help you with whatever was needed. The comfortably-sized rooms were very nice and clean, and the bathrooms were exceptionally clean and well equipped. Breakfast was continental and included in the price, offering baguettes, croissants, cereal, yogurts, etc.
 \\ 
 \hline  
\textit{Random \newline Review} & 
A real parisienne experience. The hotel is a beautiful building and the location perfect. We walked everywhere and the local area is a delight with many bars and FRIENDLY restaurants. We are an older couple and went to Paris to get engaged - it was all perfect and the hotel's ambience added to the experience. Highly recommended if you want a small, local hotel as opposed to a large modern one. Many thanks - we would return and recommend you to friends. Margaret and David
 \\ 
 \hline  
\textit{SemAE} & 
We were able to walk to all the tourists sights, except Montmarte and the Eiffel Tower.	The hotel staff was very friendly and helpful during our stay.	The staff is very friendly and helpful and the breakfast is a perfect start to the day.	The rooms was sufficent in size, the bed and pillows were very comfortable.	The ladies in the breakfast area were very friendly and courteous.	The bathroom was clean and the shower
 \\ 
 \hline  
\textit{CopyCat} & 
We stayed at this hotel for 3 nights. The hotel was very clean and the staff very helpful. The location is perfect for exploring the city and the metro station is a short walk away. We will be back!
 \\ 
 \hline  
\textit{COOP} & 
The hotel is in a great location, close to the metro and the metro. The staff were very friendly and helpful. The rooms were clean and comfortable, and the breakfast was very good. The hotel is in a great location, close to the Metro and the metro. The only downside is that the rooms are a little on the small side, but they are very clean and the staff is very friendly and helpful.
 \\ 
 \hline  
\textit{Hercules\textsubscript{abs}} & 
The room was clean and comfortable. The staff was very friendly and helpful. walking distance to everything. The room was clean and comfortable. breakfast was good. The hotel is in a great location, just a few minutes walk from the train station. breakfast was fine. The room and bathroom were very clean. The staff spoke English and were very helpful. There is also a small restaurant on the ground floor.
 \\ 
 \hline  
\textit{Mistral 7B} & 
The St. Christophe Hotel in Paris offers small, clean, and cozy rooms in a lively neighborhood near the Place Contrascarpe and Rue Mouffetard. The staff are helpful and friendly, and the location is convenient with easy access to metro stations and nearby restaurants. The hotel lacks air conditioning, so it may not be ideal during warm weather. Free Wi-Fi is available in the lobby. Overall, the St. Christophe Hotel is a charming and elegant choice for travelers to Paris.
 \\ 
 \hline  
\textit{HIRO\textsubscript{ext}} & 
Nearby, and very close to 2 Metro stations. The hotel is 1 block to the Metro (line 7) between the North and South exits of the "place Monge" stop. The bathroom was clean and functional. The staff were friendly and helpful at all times. Breakfast was Plentiful-Cold cereal, great coffee, yogurt and a wonderful basket of Croissants, breads and Pastries. The shower excellent.
 \\ 
 \hline  
\textit{HIRO\textsubscript{sent}} & 
The Hotel Saint Christophe is conveniently located near multiple metro stations. The Hotel Saint Christophe is centrally located near the Metro and restaurants in the Latin Quarter. The Hotel Saint Christophe offers clean, functional bathrooms in small rooms. The Hotel Saint Christophe's friendly and helpful staff provided excellent service. Breakfast offers cereal, yogurt, pastries, and hot beverages. Guests praised the Hotel Saint Christophe for its excellent showers.
 \\ 
 \hline  
\textit{HIRO\textsubscript{doc}} & 
The Hotel Saint Christophe is centrally located in the Latin Quarter, close to multiple Metro stations and attractions such as Notre Dame and Jardin Luxembourg. Rooms are clean and adequate, though small, and breakfast offers a variety of options. The staff are friendly and helpful, making for a pleasant stay. The hotel is conveniently located near shops, restaurants, and markets. The bathrooms are clean and modern, though some may be smaller than expected for Paris.
    \end{tabular}
    \caption{Examples of generated summaries for The Hotel Saint Christophe from \textsc{Space}. HIRO\textsubscript{doc} generates concise, informative and coherent summaries. Mistral 7B prompted directly with reviews also generates coherent summaries, but is limited to a small sample of reviews which do not necessarily reflect the overall balance of opinions. Previous abstractive systems, CopyCat, COOP and Hercules\textsubscript{abs}, generate generic summaries that lack detail.}
    \label{tab:output}
\end{table*}

\begin{table*}[ht]
\renewcommand{\arraystretch}{1.2}
    \centering
    \small
    \begin{tabular}{@{}m{2cm}@{~}|@{~}m{13.2cm}@{}}
    \textbf{System} & \textbf{Output} \\
    \hline\hline
\textit{Reference} & 
These headphones are your best option if you're looking for noise-cancelling earbuds, but they are pricey. Noise-cancelling earbuds. Available in black or white. Stay Hear tips keep earbuds in your ears while maximizing comfort. Impressive sound quality. Carrying case included. These are expensive for earbuds, so you should only buy them if you're going to use them often
 \\ 
 \hline  
\textit{Random \newline Review} & 
Excellent. I’m comparing this to the Bose on ear headphones and this cancels noise better. I’m a frequent flyer and like to disappear during flights. This helps me get into another zone whether it’s listening to electro-house or watching the latest movies while flying. Highly recommend it!
 \\ 
 \hline  
\textit{CopyCat} & 
So far so good. The only issue I have is that the volume is not as loud as my old one, but other than that I'm happy with this purchase.
 \\ 
 \hline  
\textit{COOP} & 
These are the best ear buds out of the box. The sound quality is great, the bass is very good, and the noise cancelling feature is amazing. The only downside is that it doesn't have a USB-C port on the back of the device. It's not a big deal, but if you want to use it for work, you can't go wrong with this product.
 \\ 
 \hline  
\textit{Hercules\textsubscript{abs}} & 
Love these Earbuds. These are great for listening to music. Noise cancelling works well. Long battery life. These are my favorite Headphones. The sound quality is very clear. The sound quality is great and the noise cancellation is good. Got this for my wife for Christmas. The sound quality is great and the Mic is clear. The Bass is a bit lacking. The ear pads are soft and comfortable. These Headphones are the best ever. The Headphones are very comfortable.
 \\ 
 \hline  
\textit{Mistral 7B} & 
These earbuds have received positive reviews from users who travel frequently, find them a game changer, and have even convinced friends to buy them. Some users have purchased them for their children who are also satisfied. The earbuds have received five-star ratings from both the buyers and their sons. One user's wife loves them and uses them when the user snores at night. However, some users have noted that the noise-cancelling feature is not as effective as over-ear headphones.
 \\ 
 \hline  
\textit{HIRO\textsubscript{ext}} & 
The noise cancelling is fantastic. These are the best Headphones I've ever owned. Well worth the money! The batteries last a long time and charge quickly. I have the over the ears pair and also these in the ears.
 \\ 
 \hline  
\textit{HIRO\textsubscript{sent}} & 
The Bose QuietComfort 20 headphones offer exceptional noise cancellation and comfort. These Bose QuietComfort 20 headphones are exceptional, providing excellent noise cancellation and sound quality. These headphones are worth the investment due to their exceptional noise-cancelling capabilities. Battery lasts long for noise cancellation, charging quickly. Some users report over 10 hours usage. Some prefer Bose QuietComfort 20 earbuds for flights and noisy environments, others find them dizzying.
 \\ 
 \hline  
\textit{HIRO\textsubscript{doc}} & 
The Bose QuietComfort 20 Acoustic Noise Cancelling Headphones are widely praised for their excellent noise cancellation capability, making them a popular choice for frequent flyers and those working in noisy environments. They are also praised for their comfort, sound quality, and long battery life. However, some customers have expressed concerns about the non-replaceable battery and the price. Overall, these headphones are considered a worthwhile investment for their impressive noise cancellation and sound quality.
    \end{tabular}
    \caption{Examples of generated summaries for the `Bose QuietComfort 20' headphones, from AmaSum. Mistral 7B refers to the opinions of a single user, which is not appropriate for a summary of thousands of reviews.}
    \vspace*{-.2cm}
    \label{tab:output2}
\end{table*}

\begin{table*}[ht]
    
\renewcommand{\arraystretch}{1.2}
    \centering
    \small
    \begin{tabular}{@{}m{2cm}@{~}|@{~}m{13.2cm}@{}}
    \hline\hline
    \textit{$q_{1:d}$} & (6,2,10) \\
    \hline
\multirow{16}{*}{\textit{Evidence}} & The pool area was very nice. \\ 
 & The staff was very Friendly and helpful. \\ 
 & the pool area was very enjoyable \\ 
 & The pool area is very nice. \\ 
 & The pool area is fantastic \\ 
 & The pool area was nice \\ 
 & the pool area is very attractive \\ 
 & The staff was so friendly and helpful. \\ 
 & The pool was nice \\ 
 & The staff was friendly and accommodating. \\ 
 & The pool was very nice and the lobby inviting \\ 
 & the staff was welcoming and helpful \\ 
 & The hotel staff was very friendly and accommodating. \\ 
 & Staff was friendly and helpful. \\ 
 & The pool area is very nice and large with several water features. \\ 
 & The staff was really great and helpful \\ 
\hline
\textit{HIRO\textsubscript{sent}}  &  The pool area and friendly staff make this hotel a enjoyable stay. \\

    \hline\hline
    \textit{$q_{1:d}$} & (10, 8, 6) \\
    \hline
\multirow{6}{*}{\textit{Evidence}} & 
Free parking is also available to the rear of the hotel. \\ 
 & Free parking in the hotel yard. \\ \ 
 & A big plus was the free parking and large selection for continental breakfast. \\ 
 & The free plentiful parking (some even under cover) is also one of the better features of the hotel. \\ 
 & It has free parking (a big deal in Key West), an awesome pool, and the best free continental breakfast I've ever seen. \\ 
 & Parking is free with some covered parking. \\
 
\hline
\textit{HIRO\textsubscript{sent}} &  Free parking and continental breakfast are notable features of this hotel. \\

    \hline\hline
    \textit{$q_{1:d}$} & (9,) \\
    \hline
\multirow{9}{*}{\textit{Evidence}} & I was extremely disappointed in the rooms \\ 
 & Only one mirror in the room. \\ 
 & The rooms are loud. \\ 
 & the rooms are awful \\ 
 & the room was small and shabby \\ 
 & we were a little disappointed because the room was a lot smaller than we expected \\ 
 & Extremely disappointed in the room, although the help was very nice as was the outdoor area. \\ 
 & i suspect we were in the older part of the hotel with a double room. \\ 
 & Since the parking is directly under the rooms, it was Very loud espcially from 12-3 am. \\ 
\hline
\textit{HIRO\textsubscript{sent}}   &  Rooms were small, loud, and in need of renovation with poor housekeeping. \\
\hline\hline

\textit{HIRO\textsubscript{doc}}   & The Fairfield Inn and Suites Key West received positive reviews for its friendly and helpful staff, attractive and nice pool area, and free parking. However, some guests were disappointed with the small and shabby rooms, lack of storage space, and noise from parking and adjacent rooms. The continental breakfast was also mentioned as a plus. \\
\hline\hline
    
    \end{tabular}
    \caption{Examples of selected subpaths $q_{1:d}$, the corresponding evidential clusters, the resulting HIRO\textsubscript{sent} output sentences, and the overall HIRO\textsubscript{doc} summary for a single entity from \textsc{Space}. We show only three out of five input clusters, and a subset of all evidence sentences, due to space constraints.}
    \vspace*{-.2cm}
    \label{tab:evidence_example}
\end{table*}

%% file: main.bbl
\begin{thebibliography}{63}
\expandafter\ifx\csname natexlab\endcsname\relax\def\natexlab#1{#1}\fi

\bibitem[{Aharoni et~al.(2023)Aharoni, Narayan, Maynez, Herzig, Clark, and Lapata}]{aharoni-etal-2023-multilingual}
Roee Aharoni, Shashi Narayan, Joshua Maynez, Jonathan Herzig, Elizabeth Clark, and Mirella Lapata. 2023.
\newblock \href {https://doi.org/10.18653/v1/2023.findings-acl.220} {Multilingual summarization with factual consistency evaluation}.
\newblock In \emph{Findings of the Association for Computational Linguistics: ACL 2023}, pages 3562--3591, Toronto, Canada. Association for Computational Linguistics.

\bibitem[{Amplayo et~al.(2021{\natexlab{a}})Amplayo, Angelidis, and Lapata}]{amplayo-etal-2021-aspect}
Reinald~Kim Amplayo, Stefanos Angelidis, and Mirella Lapata. 2021{\natexlab{a}}.
\newblock \href {https://doi.org/10.18653/v1/2021.emnlp-main.528} {Aspect-controllable opinion summarization}.
\newblock In \emph{Proceedings of the 2021 Conference on Empirical Methods in Natural Language Processing}, pages 6578--6593, Online and Punta Cana, Dominican Republic. Association for Computational Linguistics.

\bibitem[{Amplayo et~al.(2021{\natexlab{b}})Amplayo, Angelidis, and Lapata}]{amplayo2021unsupervised}
Reinald~Kim Amplayo, Stefanos Angelidis, and Mirella Lapata. 2021{\natexlab{b}}.
\newblock \href {https://ojs.aaai.org/index.php/AAAI/article/view/17481} {Unsupervised opinion summarization with content planning}.
\newblock In \emph{Thirty-Fifth {AAAI} Conference on Artificial Intelligence, {AAAI} 2021, Thirty-Third Conference on Innovative Applications of Artificial Intelligence, {IAAI} 2021, The Eleventh Symposium on Educational Advances in Artificial Intelligence, {EAAI} 2021, Virtual Event, February 2-9, 2021}, pages 12489--12497. {AAAI} Press.

\bibitem[{Angelidis et~al.(2021)Angelidis, Amplayo, Suhara, Wang, and Lapata}]{angelidis-etal-2021-extractive}
Stefanos Angelidis, Reinald~Kim Amplayo, Yoshihiko Suhara, Xiaolan Wang, and Mirella Lapata. 2021.
\newblock \href {https://doi.org/10.1162/tacl_a_00366} {Extractive opinion summarization in quantized transformer spaces}.
\newblock \emph{Transactions of the Association for Computational Linguistics}, 9:277--293.

\bibitem[{Basu Roy~Chowdhury et~al.(2022)Basu Roy~Chowdhury, Zhao, and Chaturvedi}]{basu-roy-chowdhury-etal-2022-unsupervised}
Somnath Basu Roy~Chowdhury, Chao Zhao, and Snigdha Chaturvedi. 2022.
\newblock \href {https://doi.org/10.18653/v1/2022.acl-long.86} {Unsupervised extractive opinion summarization using sparse coding}.
\newblock In \emph{Proceedings of the 60th Annual Meeting of the Association for Computational Linguistics (Volume 1: Long Papers)}, pages 1209--1225, Dublin, Ireland. Association for Computational Linguistics.

\bibitem[{Beltagy et~al.(2020)Beltagy, Peters, and Cohan}]{Beltagy2020Longformer}
Iz~Beltagy, Matthew~E. Peters, and Arman Cohan. 2020.
\newblock Longformer: The long-document transformer.
\newblock \emph{arXiv:2004.05150}.

\bibitem[{Bhaskar et~al.(2023)Bhaskar, Fabbri, and Durrett}]{bhaskar-etal-2023-prompted}
Adithya Bhaskar, Alex Fabbri, and Greg Durrett. 2023.
\newblock \href {https://doi.org/10.18653/v1/2023.findings-acl.591} {Prompted opinion summarization with {GPT}-3.5}.
\newblock In \emph{Findings of the Association for Computational Linguistics: ACL 2023}, pages 9282--9300, Toronto, Canada. Association for Computational Linguistics.

\bibitem[{Bra{\v{z}}inskas et~al.(2020)Bra{\v{z}}inskas, Lapata, and Titov}]{brazinskas-etal-2020-unsupervised}
Arthur Bra{\v{z}}inskas, Mirella Lapata, and Ivan Titov. 2020.
\newblock \href {https://doi.org/10.18653/v1/2020.acl-main.461} {Unsupervised opinion summarization as copycat-review generation}.
\newblock In \emph{Proceedings of the 58th Annual Meeting of the Association for Computational Linguistics}, pages 5151--5169, Online. Association for Computational Linguistics.

\bibitem[{Bra{\v{z}}inskas et~al.(2021)Bra{\v{z}}inskas, Lapata, and Titov}]{brazinskas-etal-2021-learning}
Arthur Bra{\v{z}}inskas, Mirella Lapata, and Ivan Titov. 2021.
\newblock \href {https://doi.org/10.18653/v1/2021.emnlp-main.743} {Learning opinion summarizers by selecting informative reviews}.
\newblock In \emph{Proceedings of the 2021 Conference on Empirical Methods in Natural Language Processing}, pages 9424--9442, Online and Punta Cana, Dominican Republic. Association for Computational Linguistics.

\bibitem[{Callison-Burch et~al.(2006)Callison-Burch, Osborne, and Koehn}]{callison-burch-etal-2006-evaluating}
Chris Callison-Burch, Miles Osborne, and Philipp Koehn. 2006.
\newblock \href {https://aclanthology.org/E06-1032} {Re-evaluating the role of {B}leu in machine translation research}.
\newblock In \emph{11th Conference of the {E}uropean Chapter of the Association for Computational Linguistics}, pages 249--256, Trento, Italy. Association for Computational Linguistics.

\bibitem[{Cattan et~al.(2023)Cattan, Eden, Kantor, and Bar-Haim}]{cattan-etal-2023-key}
Arie Cattan, Lilach Eden, Yoav Kantor, and Roy Bar-Haim. 2023.
\newblock \href {https://doi.org/10.18653/v1/2023.acl-long.52} {From key points to key point hierarchy: Structured and expressive opinion summarization}.
\newblock In \emph{Proceedings of the 61st Annual Meeting of the Association for Computational Linguistics (Volume 1: Long Papers)}, pages 912--928, Toronto, Canada. Association for Computational Linguistics.

\bibitem[{Chen et~al.(2010)Chen, Tao, and Wang}]{rvq}
Yongjian Chen, Guan Tao, and Cheng Wang. 2010.
\newblock \href {https://doi.org/10.3390/s101211259} {Approximate nearest neighbor search by residual vector quantization}.
\newblock \emph{Sensors (Basel, Switzerland)}, 10:11259--73.

\bibitem[{Clark et~al.(2023)Clark, Rijhwani, Gehrmann, Maynez, Aharoni, Nikolaev, Sellam, Siddhant, Das, and Parikh}]{clark-etal-2023-seahorse}
Elizabeth Clark, Shruti Rijhwani, Sebastian Gehrmann, Joshua Maynez, Roee Aharoni, Vitaly Nikolaev, Thibault Sellam, Aditya Siddhant, Dipanjan Das, and Ankur Parikh. 2023.
\newblock \href {https://doi.org/10.18653/v1/2023.emnlp-main.584} {{SEAHORSE}: A multilingual, multifaceted dataset for summarization evaluation}.
\newblock In \emph{Proceedings of the 2023 Conference on Empirical Methods in Natural Language Processing}, pages 9397--9413, Singapore. Association for Computational Linguistics.

\bibitem[{Coavoux et~al.(2019)Coavoux, Elsahar, and Gall{\'e}}]{coavoux-etal-2019-unsupervised}
Maximin Coavoux, Hady Elsahar, and Matthias Gall{\'e}. 2019.
\newblock \href {https://doi.org/10.18653/v1/D19-5405} {Unsupervised aspect-based multi-document abstractive summarization}.
\newblock In \emph{Proceedings of the 2nd Workshop on New Frontiers in Summarization}, pages 42--47, Hong Kong, China. Association for Computational Linguistics.

\bibitem[{Devlin et~al.(2019)Devlin, Chang, Lee, and Toutanova}]{devlin-etal-2019-bert}
Jacob Devlin, Ming-Wei Chang, Kenton Lee, and Kristina Toutanova. 2019.
\newblock \href {https://doi.org/10.18653/v1/N19-1423} {{BERT}: Pre-training of deep bidirectional transformers for language understanding}.
\newblock In \emph{Proceedings of the 2019 Conference of the North {A}merican Chapter of the Association for Computational Linguistics: Human Language Technologies, Volume 1 (Long and Short Papers)}, pages 4171--4186, Minneapolis, Minnesota. Association for Computational Linguistics.

\bibitem[{Erkan and Radev(2004)}]{lexrank}
G.~Erkan and D.~R. Radev. 2004.
\newblock \href {https://doi.org/10.1613/jair.1523} {{LexRank}: Graph-based lexical centrality as salience in text summarization}.
\newblock \emph{Journal of Artificial Intelligence Research}, 22:457--479.

\bibitem[{Fabbri et~al.(2021)Fabbri, Kryściński, McCann, Xiong, Socher, and Radev}]{10.1162/tacl_a_00373}
Alexander~R. Fabbri, Wojciech Kryściński, Bryan McCann, Caiming Xiong, Richard Socher, and Dragomir Radev. 2021.
\newblock \href {https://doi.org/10.1162/tacl_a_00373} {{SummEval: Re-evaluating Summarization Evaluation}}.
\newblock \emph{Transactions of the Association for Computational Linguistics}, 9:391--409.

\bibitem[{Goyal et~al.(2022)Goyal, Li, and Durrett}]{goyal2022zeroshotnews}
Tanya Goyal, Junyi~Jessy Li, and Greg Durrett. 2022.
\newblock News summarization and evaluation in the era of {GPT}-3.
\newblock \emph{arXiv preprint}.

\bibitem[{Gu et~al.(2022)Gu, Goel, and R{\'{e}}}]{s4}
Albert Gu, Karan Goel, and Christopher R{\'{e}}. 2022.
\newblock \href {https://openreview.net/forum?id=uYLFoz1vlAC} {Efficiently modeling long sequences with structured state spaces}.
\newblock In \emph{The Tenth International Conference on Learning Representations, {ICLR} 2022, Virtual Event, April 25-29, 2022}. OpenReview.net.

\bibitem[{He et~al.(2021)He, Gao, and Chen}]{he2021debertav3}
Pengcheng He, Jianfeng Gao, and Weizhu Chen. 2021.
\newblock \href {http://arxiv.org/abs/2111.09543} {Debertav3: Improving deberta using electra-style pre-training with gradient-disentangled embedding sharing}.

\bibitem[{Hofstätter et~al.(2021)Hofstätter, Althammer, Schröder, Sertkan, and Hanbury}]{hofstatter2021improving}
Sebastian Hofstätter, Sophia Althammer, Michael Schröder, Mete Sertkan, and Allan Hanbury. 2021.
\newblock \href {http://arxiv.org/abs/2010.02666} {Improving efficient neural ranking models with cross-architecture knowledge distillation}.

\bibitem[{Hosking et~al.(2023{\natexlab{a}})Hosking, Blunsom, and Bartolo}]{hosking2023human}
Tom Hosking, Phil Blunsom, and Max Bartolo. 2023{\natexlab{a}}.
\newblock \href {http://arxiv.org/abs/2309.16349} {Human feedback is not gold standard}.

\bibitem[{Hosking et~al.(2023{\natexlab{b}})Hosking, Tang, and Lapata}]{hosking-etal-2023-attributable}
Tom Hosking, Hao Tang, and Mirella Lapata. 2023{\natexlab{b}}.
\newblock \href {https://doi.org/10.18653/v1/2023.acl-long.473} {Attributable and scalable opinion summarization}.
\newblock In \emph{Proceedings of the 61st Annual Meeting of the Association for Computational Linguistics (Volume 1: Long Papers)}, pages 8488--8505, Toronto, Canada. Association for Computational Linguistics.

\bibitem[{Hu and Liu(2004)}]{10.1145/1014052.1014073}
Minqing Hu and Bing Liu. 2004.
\newblock \href {https://doi.org/10.1145/1014052.1014073} {Mining and summarizing customer reviews}.
\newblock In \emph{Proceedings of the Tenth ACM SIGKDD International Conference on Knowledge Discovery and Data Mining}, KDD '04, page 168–177, New York, NY, USA. Association for Computing Machinery.

\bibitem[{Hubert and Arabie(1985)}]{ari}
Lawrence Hubert and Phipps Arabie. 1985.
\newblock \href {https://EconPapers.repec.org/RePEc:spr:jclass:v:2:y:1985:i:1:p:193-218} {Comparing partitions}.
\newblock \emph{Journal of Classification}, 2(1):193--218.

\bibitem[{Iso et~al.(2021)Iso, Wang, Suhara, Angelidis, and Tan}]{iso-etal-2021-convex-aggregation}
Hayate Iso, Xiaolan Wang, Yoshihiko Suhara, Stefanos Angelidis, and Wang-Chiew Tan. 2021.
\newblock \href {https://doi.org/10.18653/v1/2021.findings-emnlp.328} {{C}onvex {A}ggregation for {O}pinion {S}ummarization}.
\newblock In \emph{Findings of the Association for Computational Linguistics: EMNLP 2021}, pages 3885--3903, Punta Cana, Dominican Republic. Association for Computational Linguistics.

\bibitem[{Jang et~al.(2017)Jang, Gu, and Poole}]{jang2016categorical}
Eric Jang, Shixiang Gu, and Ben Poole. 2017.
\newblock \href {https://openreview.net/forum?id=rkE3y85ee} {Categorical reparameterization with {G}umbel-softmax}.
\newblock In \emph{5th International Conference on Learning Representations, {ICLR} 2017, Toulon, France, April 24-26, 2017, Conference Track Proceedings}. OpenReview.net.

\bibitem[{Jiang et~al.(2023)Jiang, Sablayrolles, Mensch, Bamford, Chaplot, de~las Casas, Bressand, Lengyel, Lample, Saulnier, Lavaud, Lachaux, Stock, Scao, Lavril, Wang, Lacroix, and Sayed}]{jiang2023mistral}
Albert~Q. Jiang, Alexandre Sablayrolles, Arthur Mensch, Chris Bamford, Devendra~Singh Chaplot, Diego de~las Casas, Florian Bressand, Gianna Lengyel, Guillaume Lample, Lucile Saulnier, Lélio~Renard Lavaud, Marie-Anne Lachaux, Pierre Stock, Teven~Le Scao, Thibaut Lavril, Thomas Wang, Timothée Lacroix, and William~El Sayed. 2023.
\newblock \href {http://arxiv.org/abs/2310.06825} {Mistral 7b}.

\bibitem[{Jones(1972)}]{tfidf}
Karen~Spärck Jones. 1972.
\newblock A statistical interpretation of term specificity and its application in retrieval.
\newblock \emph{Journal of Documentation}, 28:11--21.

\bibitem[{Kedzie et~al.(2018)Kedzie, McKeown, and Daum{\'e}~III}]{kedzie-etal-2018-content}
Chris Kedzie, Kathleen McKeown, and Hal Daum{\'e}~III. 2018.
\newblock \href {https://doi.org/10.18653/v1/D18-1208} {Content selection in deep learning models of summarization}.
\newblock In \emph{Proceedings of the 2018 Conference on Empirical Methods in Natural Language Processing}, pages 1818--1828, Brussels, Belgium. Association for Computational Linguistics.

\bibitem[{Kingma and Ba(2015)}]{adam}
Diederik~P. Kingma and Jimmy Ba. 2015.
\newblock \href {http://arxiv.org/abs/1412.6980} {Adam: {A} method for stochastic optimization}.
\newblock In \emph{3rd International Conference on Learning Representations, {ICLR} 2015, San Diego, CA, USA, May 7-9, 2015, Conference Track Proceedings}.

\bibitem[{Laban et~al.(2022)Laban, Schnabel, Bennett, and Hearst}]{laban-etal-2022-summac}
Philippe Laban, Tobias Schnabel, Paul~N. Bennett, and Marti~A. Hearst. 2022.
\newblock \href {https://doi.org/10.1162/tacl_a_00453} {{S}umma{C}: Re-visiting {NLI}-based models for inconsistency detection in summarization}.
\newblock \emph{Transactions of the Association for Computational Linguistics}, 10:163--177.

\bibitem[{Lan et~al.(2020)Lan, Chen, Goodman, Gimpel, Sharma, and Soricut}]{albert}
Zhenzhong Lan, Mingda Chen, Sebastian Goodman, Kevin Gimpel, Piyush Sharma, and Radu Soricut. 2020.
\newblock \href {https://openreview.net/forum?id=H1eA7AEtvS} {{ALBERT:} {A} lite {BERT} for self-supervised learning of language representations}.
\newblock In \emph{8th International Conference on Learning Representations, {ICLR} 2020, Addis Ababa, Ethiopia, April 26-30, 2020}. OpenReview.net.

\bibitem[{Lewis et~al.(2020)Lewis, Perez, Piktus, Petroni, Karpukhin, Goyal, K\"{u}ttler, Lewis, Yih, Rockt\"{a}schel, Riedel, and Kiela}]{rag}
Patrick Lewis, Ethan Perez, Aleksandra Piktus, Fabio Petroni, Vladimir Karpukhin, Naman Goyal, Heinrich K\"{u}ttler, Mike Lewis, Wen-tau Yih, Tim Rockt\"{a}schel, Sebastian Riedel, and Douwe Kiela. 2020.
\newblock Retrieval-augmented generation for knowledge-intensive nlp tasks.
\newblock In \emph{Proceedings of the 34th International Conference on Neural Information Processing Systems}, NIPS'20, Red Hook, NY, USA. Curran Associates Inc.

\bibitem[{Li et~al.(2023)Li, Ai, Zhan, Mao, Liu, Liu, and Cao}]{10.1145/3539618.3591651}
Haitao Li, Qingyao Ai, Jingtao Zhan, Jiaxin Mao, Yiqun Liu, Zheng Liu, and Zhao Cao. 2023.
\newblock \href {https://doi.org/10.1145/3539618.3591651} {Constructing tree-based index for efficient and effective dense retrieval}.
\newblock In \emph{Proceedings of the 46th International ACM SIGIR Conference on Research and Development in Information Retrieval}, SIGIR '23, page 131–140, New York, NY, USA. Association for Computing Machinery.

\bibitem[{Lin(2004)}]{lin-2004-rouge}
Chin-Yew Lin. 2004.
\newblock \href {https://aclanthology.org/W04-1013} {{ROUGE}: A package for automatic evaluation of summaries}.
\newblock In \emph{Text Summarization Branches Out}, pages 74--81, Barcelona, Spain. Association for Computational Linguistics.

\bibitem[{Liu et~al.(2023)Liu, Lin, Hewitt, Paranjape, Bevilacqua, Petroni, and Liang}]{liu2023lost}
Nelson~F. Liu, Kevin Lin, John Hewitt, Ashwin Paranjape, Michele Bevilacqua, Fabio Petroni, and Percy Liang. 2023.
\newblock \href {http://arxiv.org/abs/2307.03172} {Lost in the middle: How language models use long contexts}.

\bibitem[{Liu and Lapata(2019)}]{liu-lapata-2019-hierarchical}
Yang Liu and Mirella Lapata. 2019.
\newblock \href {https://doi.org/10.18653/v1/P19-1500} {Hierarchical transformers for multi-document summarization}.
\newblock In \emph{Proceedings of the 57th Annual Meeting of the Association for Computational Linguistics}, pages 5070--5081, Florence, Italy. Association for Computational Linguistics.

\bibitem[{Louis and Maynez(2023)}]{louis-maynez-2023-opinesum}
Annie Louis and Joshua Maynez. 2023.
\newblock \href {https://doi.org/10.18653/v1/2023.findings-acl.686} {{O}pine{S}um: Entailment-based self-training for abstractive opinion summarization}.
\newblock In \emph{Findings of the Association for Computational Linguistics: ACL 2023}, pages 10774--10790, Toronto, Canada. Association for Computational Linguistics.

\bibitem[{Louviere and Woodworth(1990)}]{louviere1990best}
Jordan~J Louviere and George~G Woodworth. 1990.
\newblock Best worst scaling: A model for largest difference judgments [working paper].
\newblock \emph{Faculty of Business}.

\bibitem[{Maddison et~al.(2017)Maddison, Mnih, and Teh}]{maddison2017concrete}
Chris~J. Maddison, Andriy Mnih, and Yee~Whye Teh. 2017.
\newblock \href {https://openreview.net/forum?id=S1jE5L5gl} {The concrete distribution: {A} continuous relaxation of discrete random variables}.
\newblock In \emph{5th International Conference on Learning Representations, {ICLR} 2017, Toulon, France, April 24-26, 2017, Conference Track Proceedings}. OpenReview.net.

\bibitem[{Malon(2023)}]{prevalence}
C.~Malon. 2023.
\newblock Automatically evaluating opinion prevalence in opinion summarization.
\newblock In \emph{The 6th Workshop on e-Commerce and NLP (KDD 2023)}.

\bibitem[{Narayan et~al.(2023)Narayan, Maynez, Amplayo, Ganchev, Louis, Huot, Sandholm, Das, and Lapata}]{narayan-etal-2023-conditional}
Shashi Narayan, Joshua Maynez, Reinald~Kim Amplayo, Kuzman Ganchev, Annie Louis, Fantine Huot, Anders Sandholm, Dipanjan Das, and Mirella Lapata. 2023.
\newblock \href {https://doi.org/10.1162/tacl_a_00583} {Conditional generation with a question-answering blueprint}.
\newblock \emph{Transactions of the Association for Computational Linguistics}, 11:974--996.

\bibitem[{Opper et~al.(2023)Opper, Prokhorov, and N}]{opper-etal-2023-strae}
Mattia Opper, Victor Prokhorov, and Siddharth N. 2023.
\newblock \href {https://doi.org/10.18653/v1/2023.emnlp-main.469} {{S}tr{AE}: Autoencoding for pre-trained embeddings using explicit structure}.
\newblock In \emph{Proceedings of the 2023 Conference on Empirical Methods in Natural Language Processing}, pages 7544--7560, Singapore. Association for Computational Linguistics.

\bibitem[{Puduppully et~al.(2019)Puduppully, Dong, and Lapata}]{DBLP:conf/aaai/Puduppully0L19}
Ratish Puduppully, Li~Dong, and Mirella Lapata. 2019.
\newblock \href {https://doi.org/10.1609/AAAI.V33I01.33016908} {Data-to-text generation with content selection and planning}.
\newblock In \emph{The Thirty-Third {AAAI} Conference on Artificial Intelligence, {AAAI} 2019, The Thirty-First Innovative Applications of Artificial Intelligence Conference, {IAAI} 2019, The Ninth {AAAI} Symposium on Educational Advances in Artificial Intelligence, {EAAI} 2019, Honolulu, Hawaii, USA, January 27 - February 1, 2019}, pages 6908--6915. {AAAI} Press.

\bibitem[{Raffel et~al.(2020)Raffel, Shazeer, Roberts, Lee, Narang, Matena, Zhou, Li, and Liu}]{t5}
Colin Raffel, Noam Shazeer, Adam Roberts, Katherine Lee, Sharan Narang, Michael Matena, Yanqi Zhou, Wei Li, and Peter~J. Liu. 2020.
\newblock \href {http://jmlr.org/papers/v21/20-074.html} {Exploring the limits of transfer learning with a unified text-to-text transformer}.
\newblock \emph{Journal of Machine Learning Research}, 21(140):1--67.

\bibitem[{Rashkin et~al.(2023)Rashkin, Nikolaev, Lamm, Aroyo, Collins, Das, Petrov, Tomar, Turc, and Reitter}]{10.1162/coli_a_00486}
Hannah Rashkin, Vitaly Nikolaev, Matthew Lamm, Lora Aroyo, Michael Collins, Dipanjan Das, Slav Petrov, Gaurav~Singh Tomar, Iulia Turc, and David Reitter. 2023.
\newblock \href {https://doi.org/10.1162/coli_a_00486} {{Measuring Attribution in Natural Language Generation Models}}.
\newblock \emph{Computational Linguistics}, 49(4):777--840.

\bibitem[{Reimers and Gurevych(2019)}]{reimers-gurevych-2019-sentence}
Nils Reimers and Iryna Gurevych. 2019.
\newblock \href {https://doi.org/10.18653/v1/D19-1410} {Sentence-{BERT}: Sentence embeddings using {S}iamese {BERT}-networks}.
\newblock In \emph{Proceedings of the 2019 Conference on Empirical Methods in Natural Language Processing and the 9th International Joint Conference on Natural Language Processing (EMNLP-IJCNLP)}, pages 3982--3992, Hong Kong, China. Association for Computational Linguistics.

\bibitem[{Sarthi et~al.(2024)Sarthi, Abdullah, Tuli, Khanna, Goldie, and Manning}]{sarthi2024raptor}
Parth Sarthi, Salman Abdullah, Aditi Tuli, Shubh Khanna, Anna Goldie, and Christopher~D. Manning. 2024.
\newblock Raptor: Recursive abstractive processing for tree-organized retrieval.
\newblock In \emph{International Conference on Learning Representations (ICLR)}.

\bibitem[{Schuster et~al.(2021)Schuster, Fisch, and Barzilay}]{schuster-etal-2021-get}
Tal Schuster, Adam Fisch, and Regina Barzilay. 2021.
\newblock \href {https://doi.org/10.18653/v1/2021.naacl-main.52} {Get your vitamin {C}! robust fact verification with contrastive evidence}.
\newblock In \emph{Proceedings of the 2021 Conference of the North American Chapter of the Association for Computational Linguistics: Human Language Technologies}, pages 624--643, Online. Association for Computational Linguistics.

\bibitem[{Shen and Wan(2023)}]{shen2023opinsummeval}
Yuchen Shen and Xiaojun Wan. 2023.
\newblock \href {http://arxiv.org/abs/2310.18122} {Opinsummeval: Revisiting automated evaluation for opinion summarization}.

\bibitem[{S{\o}nderby et~al.(2017)S{\o}nderby, Poole, and Mnih}]{sonderby2017continuous}
Casper~Kaae S{\o}nderby, Ben Poole, and Andriy Mnih. 2017.
\newblock Continuous relaxation training of discrete latent variable image models.
\newblock In \emph{Beysian DeepLearning workshop, NIPS}, volume 201.

\bibitem[{Tay et~al.(2019)Tay, Joshi, Zhang, Karimi, and Wan}]{tay-etal-2019-red}
Wenyi Tay, Aditya Joshi, Xiuzhen Zhang, Sarvnaz Karimi, and Stephen Wan. 2019.
\newblock \href {https://aclanthology.org/U19-1008} {Red-faced {ROUGE}: Examining the suitability of {ROUGE} for opinion summary evaluation}.
\newblock In \emph{Proceedings of the 17th Annual Workshop of the Australasian Language Technology Association}, pages 52--60, Sydney, Australia. Australasian Language Technology Association.

\bibitem[{Touvron et~al.(2023)Touvron, Martin, Stone, Albert, Almahairi, Babaei, Bashlykov, Batra, Bhargava, Bhosale, Bikel, Blecher, Ferrer, Chen, Cucurull, Esiobu, Fernandes, Fu, Fu, Fuller, Gao, Goswami, Goyal, Hartshorn, Hosseini, Hou, Inan, Kardas, Kerkez, Khabsa, Kloumann, Korenev, Koura, Lachaux, Lavril, Lee, Liskovich, Lu, Mao, Martinet, Mihaylov, Mishra, Molybog, Nie, Poulton, Reizenstein, Rungta, Saladi, Schelten, Silva, Smith, Subramanian, Tan, Tang, Taylor, Williams, Kuan, Xu, Yan, Zarov, Zhang, Fan, Kambadur, Narang, Rodriguez, Stojnic, Edunov, and Scialom}]{touvron2023llama}
Hugo Touvron, Louis Martin, Kevin Stone, Peter Albert, Amjad Almahairi, Yasmine Babaei, Nikolay Bashlykov, Soumya Batra, Prajjwal Bhargava, Shruti Bhosale, Dan Bikel, Lukas Blecher, Cristian~Canton Ferrer, Moya Chen, Guillem Cucurull, David Esiobu, Jude Fernandes, Jeremy Fu, Wenyin Fu, Brian Fuller, Cynthia Gao, Vedanuj Goswami, Naman Goyal, Anthony Hartshorn, Saghar Hosseini, Rui Hou, Hakan Inan, Marcin Kardas, Viktor Kerkez, Madian Khabsa, Isabel Kloumann, Artem Korenev, Punit~Singh Koura, Marie-Anne Lachaux, Thibaut Lavril, Jenya Lee, Diana Liskovich, Yinghai Lu, Yuning Mao, Xavier Martinet, Todor Mihaylov, Pushkar Mishra, Igor Molybog, Yixin Nie, Andrew Poulton, Jeremy Reizenstein, Rashi Rungta, Kalyan Saladi, Alan Schelten, Ruan Silva, Eric~Michael Smith, Ranjan Subramanian, Xiaoqing~Ellen Tan, Binh Tang, Ross Taylor, Adina Williams, Jian~Xiang Kuan, Puxin Xu, Zheng Yan, Iliyan Zarov, Yuchen Zhang, Angela Fan, Melanie Kambadur, Sharan Narang, Aurelien Rodriguez, Robert Stojnic, Sergey Edunov, and Thomas
  Scialom. 2023.
\newblock \href {http://arxiv.org/abs/2307.09288} {Llama 2: Open foundation and fine-tuned chat models}.

\bibitem[{van~den Oord et~al.(2018)van~den Oord, Li, and Vinyals}]{infonce}
A{\"{a}}ron van~den Oord, Yazhe Li, and Oriol Vinyals. 2018.
\newblock \href {http://arxiv.org/abs/1807.03748} {Representation learning with contrastive predictive coding}.
\newblock \emph{CoRR}, abs/1807.03748.

\bibitem[{Vaswani et~al.(2017)Vaswani, Shazeer, Parmar, Uszkoreit, Jones, Gomez, Kaiser, and Polosukhin}]{Vaswani2017}
Ashish Vaswani, Noam Shazeer, Niki Parmar, Jakob Uszkoreit, Llion Jones, Aidan~N. Gomez, Lukasz Kaiser, and Illia Polosukhin. 2017.
\newblock \href {https://proceedings.neurips.cc/paper/2017/hash/3f5ee243547dee91fbd053c1c4a845aa-Abstract.html} {Attention is all you need}.
\newblock In \emph{Advances in Neural Information Processing Systems 30: Annual Conference on Neural Information Processing Systems 2017, December 4-9, 2017, Long Beach, CA, {USA}}, pages 5998--6008.

\bibitem[{Vendrov et~al.(2016)Vendrov, Kiros, Fidler, and Urtasun}]{DBLP:journals/corr/VendrovKFU15}
Ivan Vendrov, Ryan Kiros, Sanja Fidler, and Raquel Urtasun. 2016.
\newblock \href {http://arxiv.org/abs/1511.06361} {Order-embeddings of images and language}.
\newblock In \emph{4th International Conference on Learning Representations, {ICLR} 2016, San Juan, Puerto Rico, May 2-4, 2016, Conference Track Proceedings}.

\bibitem[{Wang et~al.(2023)Wang, Araki, Jiang, Parvez, and Neubig}]{wang2023learning}
Zhiruo Wang, Jun Araki, Zhengbao Jiang, Md~Rizwan Parvez, and Graham Neubig. 2023.
\newblock \href {http://arxiv.org/abs/2311.08377} {Learning to filter context for retrieval-augmented generation}.

\bibitem[{Wu et~al.(2022)Wu, Gardner, Stenetorp, and Dasigi}]{wu-etal-2022-generating}
Yuxiang Wu, Matt Gardner, Pontus Stenetorp, and Pradeep Dasigi. 2022.
\newblock \href {https://doi.org/10.18653/v1/2022.acl-long.190} {Generating data to mitigate spurious correlations in natural language inference datasets}.
\newblock In \emph{Proceedings of the 60th Annual Meeting of the Association for Computational Linguistics (Volume 1: Long Papers)}, pages 2660--2676, Dublin, Ireland. Association for Computational Linguistics.

\bibitem[{Xu et~al.(2023)Xu, Ping, Wu, McAfee, Zhu, Liu, Subramanian, Bakhturina, Shoeybi, and Catanzaro}]{xu2023retrieval}
Peng Xu, Wei Ping, Xianchao Wu, Lawrence McAfee, Chen Zhu, Zihan Liu, Sandeep Subramanian, Evelina Bakhturina, Mohammad Shoeybi, and Bryan Catanzaro. 2023.
\newblock \href {http://arxiv.org/abs/2310.03025} {Retrieval meets long context large language models}.

\bibitem[{Zeghidour et~al.(2022)Zeghidour, Luebs, Omran, Skoglund, and Tagliasacchi}]{10.1109/TASLP.2021.3129994}
Neil Zeghidour, Alejandro Luebs, Ahmed Omran, Jan Skoglund, and Marco Tagliasacchi. 2022.
\newblock \href {https://doi.org/10.1109/TASLP.2021.3129994} {Soundstream: An end-to-end neural audio codec}.
\newblock \emph{IEEE/ACM Trans. Audio, Speech and Lang. Proc.}, 30:495–507.

\bibitem[{Zhang et~al.(2024)Zhang, Xu, and Perez-Beltrachini}]{zhang-etal-2024-fine}
Huajian Zhang, Yumo Xu, and Laura Perez-Beltrachini. 2024.
\newblock \href {https://aclanthology.org/2024.eacl-long.102} {Fine-grained natural language inference based faithfulness evaluation for diverse summarisation tasks}.
\newblock In \emph{Proceedings of the 18th Conference of the European Chapter of the Association for Computational Linguistics (Volume 1: Long Papers)}, pages 1701--1722, St. Julian{'}s, Malta. Association for Computational Linguistics.

\bibitem[{Zhao et~al.(2022)Zhao, Yang, Li, Wang, Wu, Ren, de~Rijke, and Ren}]{10.1145/3477495.3532037}
Mengxue Zhao, Yang Yang, Miao Li, Jingang Wang, Wei Wu, Pengjie Ren, Maarten de~Rijke, and Zhaochun Ren. 2022.
\newblock \href {https://doi.org/10.1145/3477495.3532037} {Personalized abstractive opinion tagging}.
\newblock In \emph{Proceedings of the 45th International ACM SIGIR Conference on Research and Development in Information Retrieval}, SIGIR '22, page 1066–1076, New York, NY, USA. Association for Computing Machinery.

\end{thebibliography}
